\documentclass{article}

\usepackage{microtype}
\usepackage{graphicx}
\usepackage{subfigure}
\usepackage{booktabs} 

\usepackage{hyperref}

\usepackage[accepted]{icml2024}

\usepackage{amsmath}
\usepackage{amssymb}
\usepackage{mathtools}
\usepackage{amsthm}

\usepackage[capitalize,noabbrev]{cleveref}

\usepackage{times}
\usepackage{latexsym}
\usepackage{graphicx}
\usepackage{amsmath}
\usepackage{lipsum}
\usepackage{multicol}
\usepackage{amssymb}
\usepackage{mathtools}
\usepackage{makecell}

\usepackage[T1]{fontenc}

\usepackage[utf8]{inputenc}

\usepackage{microtype}
\usepackage{tabularx,ragged2e}
\usepackage{float}
\usepackage{bbm}

\DeclareMathOperator*{\argmax}{\arg\max}
\usepackage{enumitem}
\usepackage{subcaption}
\usepackage[normalem]{ulem}
\useunder{\uline}{\ul}{}
\usepackage{diagbox}

\theoremstyle{plain}

\theoremstyle{definition}

\theoremstyle{remark}

\usepackage[textsize=tiny]{todonotes}

\icmltitlerunning{RLAIF vs. RLHF}

\begin{document}

\twocolumn[
\icmltitle{RLAIF vs. RLHF: Scaling Reinforcement Learning \\
from Human Feedback with AI Feedback}




\icmlsetsymbol{equal}{*}

\begin{icmlauthorlist}
\icmlauthor{Harrison Lee}{GDM}
\icmlauthor{Samrat Phatale}{GDM}
\icmlauthor{Hassan Mansoor}{GDM}
\icmlauthor{Thomas Mesnard}{GDM}
\icmlauthor{Johan Ferret}{GDM}
\icmlauthor{Kellie Lu}{G}
\icmlauthor{Colton Bishop}{GDM}
\icmlauthor{Ethan Hall}{G}
\icmlauthor{Victor Carbune}{GDM}
\icmlauthor{Abhinav Rastogi}{GDM}
\icmlauthor{Sushant Prakash}{G}
\end{icmlauthorlist}

\icmlaffiliation{GDM}{Google DeepMind}
\icmlaffiliation{G}{Google}

\icmlcorrespondingauthor{Harrison Lee}{harrisonlee@google.com}
\icmlcorrespondingauthor{Samrat Phatale}{samratph@google.com}
\icmlcorrespondingauthor{Hassan Mansoor}{hassan@google.com}

\icmlkeywords{Reinforcement Learning, Large Language Models, Alignment with Human Feedback, Alignment from AI Feedback, Self-improvement}

\vskip 0.3in
]



\printAffiliationsAndNotice{}  

\begin{abstract}
Reinforcement learning from human feedback (RLHF) has proven effective in aligning large language models (LLMs) with human preferences, but gathering high-quality preference labels is expensive. RL from AI Feedback (RLAIF), introduced in \citet{bai2022constitutional}, offers a promising alternative that trains the reward model (RM) on preferences generated by an off-the-shelf LLM. Across the tasks of summarization, helpful dialogue generation, and harmless dialogue generation, we show that RLAIF achieves comparable performance to RLHF. Furthermore, we take a step towards ``self-improvement'' by demonstrating that RLAIF can outperform a supervised fine-tuned baseline even when the AI labeler is the same size as the policy, or even the exact same checkpoint as the initial policy. Finally, we introduce direct-RLAIF (d-RLAIF) - a technique that circumvents RM training by obtaining rewards directly from an off-the-shelf LLM during RL, which achieves superior performance to canonical RLAIF. Our results suggest that RLAIF can achieve performance on-par with using human feedback, offering a potential solution to the scalability limitations of RLHF.
\end{abstract}

\begin{figure}[tb]
    \begin{center}
    
    \begin{subfigure}
        
        \includegraphics[width=0.8\columnwidth]{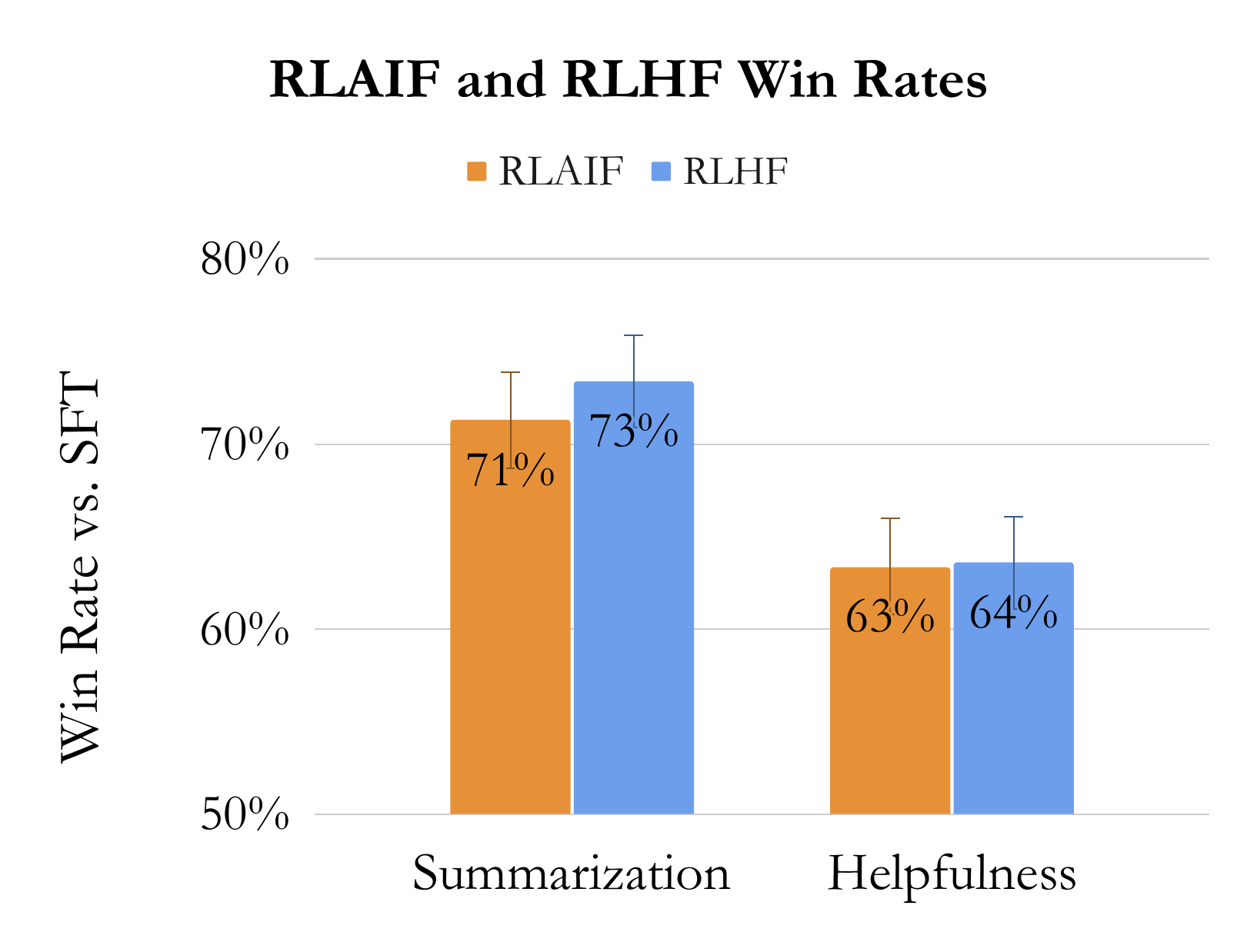}
    \end{subfigure}
    \vskip -0.1in
    \begin{subfigure}
        
        \includegraphics[width=0.8\columnwidth]{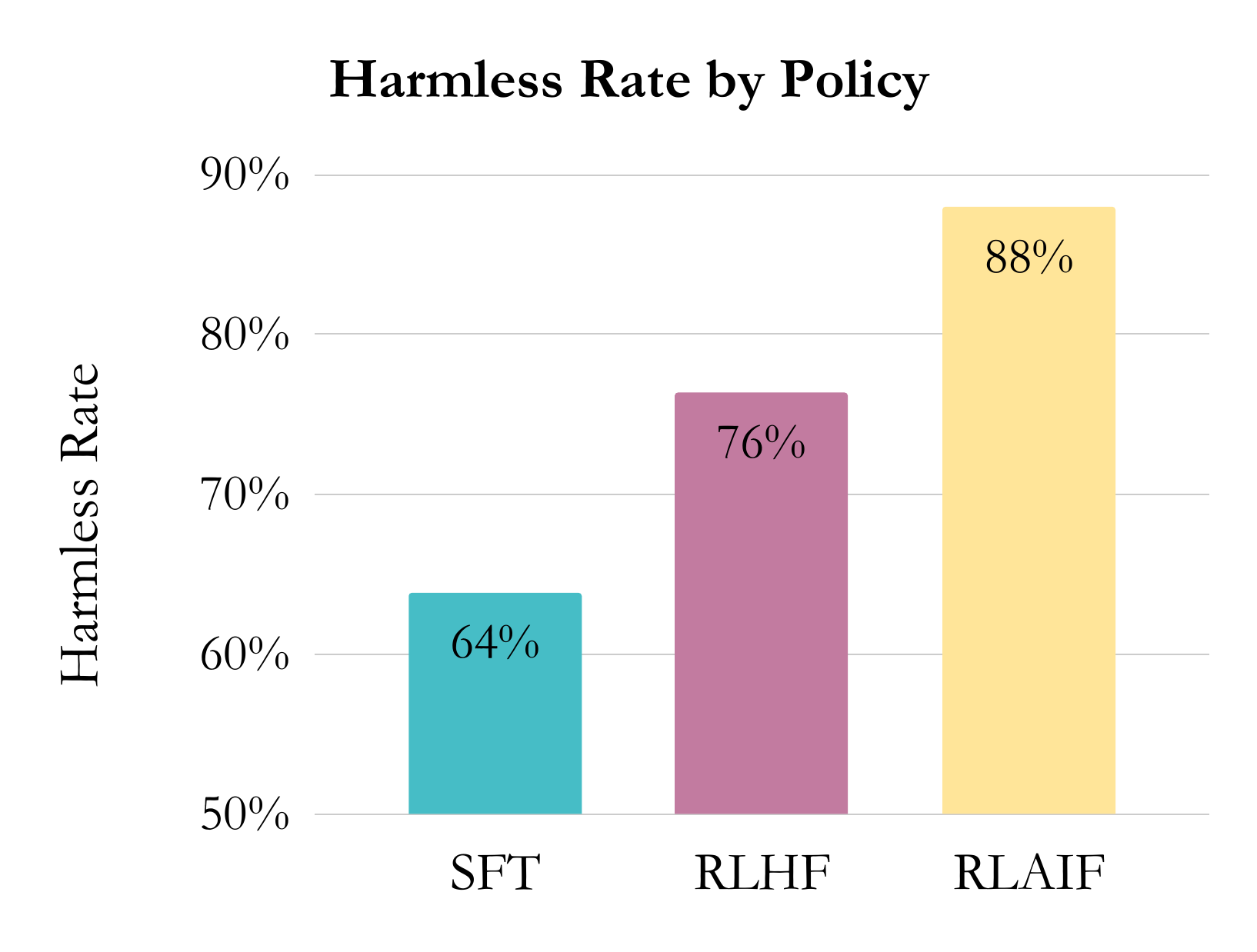}
    \end{subfigure}
    
    \vskip -0.1in 

    \caption{Human evaluators strongly prefer RLAIF and RLHF over the SFT baseline for summarization and helpful dialogue generation. Furthermore, when compared head-to-head, RLAIF is equally preferred to RLHF. For harmless dialogue generation, RLAIF outperforms RLHF.}
    \vskip -0.25in
    \label{fig:main_result}
    \end{center}
\end{figure}

\section{Introduction}

Reinforcement Learning from Human Feedback (RLHF) is an effective technique for aligning language models to human preferences \citep{stiennon2020learning,ouyang2022training}. It is cited as one of the key drivers of success in modern conversational language models, such as ChatGPT \citep{liu2023summary} and Bard \citep{bardoverview}. A key advantage of training language models with reinforcement learning (RL) is that it enables optimization on complex, sequence-level objectives that are not easily differentiable and therefore ill-suited for traditional supervised fine-tuning (SFT).

\begin{figure*}
\begin{center}
    \includegraphics[width=135mm]{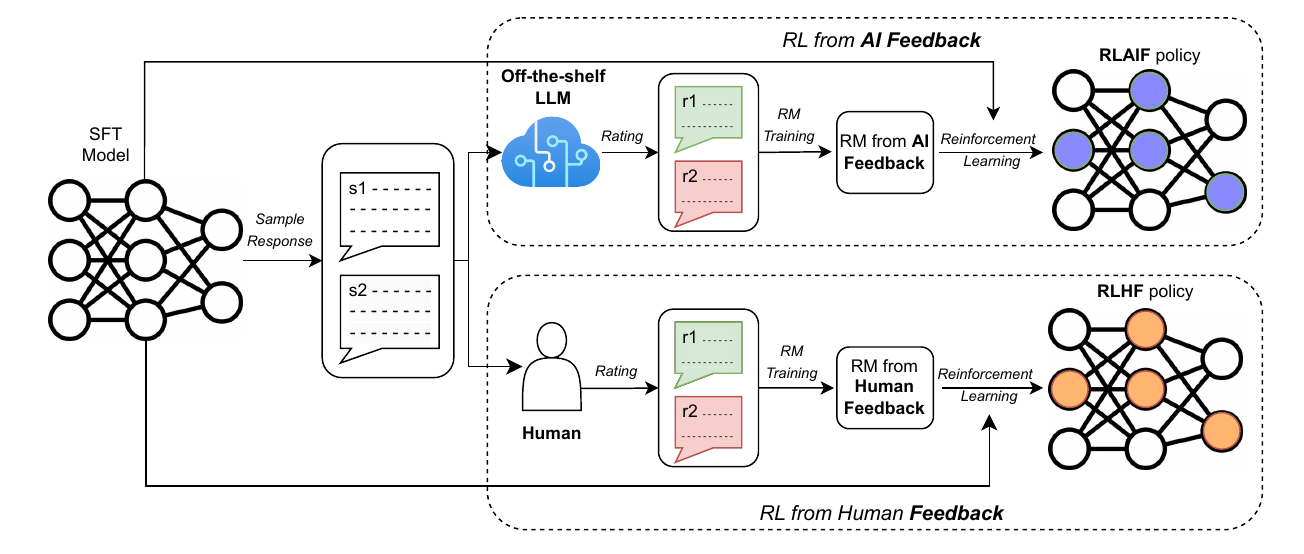}
    \vskip -0.1in
    \caption{A diagram depicting RLAIF (top) vs. RLHF (bottom)}
    \vskip -0.1in
    \label{fig:process}
\end{center}
\end{figure*}

One obstacle for employing RLHF at scale is its dependence on high-quality human preference labels. Modern large language models (LLMs) have shown a high degree of alignment with human judgment \citep{gilardi2023chatgpt,ding-etal-2023-gpt}, suggesting that LLM-generated preference labels may be a viable substitute for human labels. \citet{bai2022constitutional} was the first effort to explore Reinforcement Learning from AI Feedback (RLAIF), where RL was conducted using a reward model trained on a hybrid of human and AI preferences. In conjunction with their ``Constitutional AI'' self-revision technique, their final policy outperformed supervised fine-tuning for training a conversational assistant. However, it did not directly compare the efficacy of human vs. AI feedback, leaving the question of whether RLAIF can be a suitable alternative to RLHF unanswered.

In this work, we compare the effectiveness of RLAIF and RLHF (see Figure \ref{fig:process}) on three tasks: summarization, helpful dialogue generation, and harmless dialogue generation. Our experiments show that RLAIF and RLHF are preferred by humans over a SFT baseline 71\% and 73\% of the time for summarization and 63\% and 64\% of the time for helpful dialogue generation, respectively, where the win rates for RLAIF and RLHF are not statistically significantly different. Furthermore, in a head-to-head comparison of RLAIF against RLHF, both policies are equally preferred\footnote{The win rate for one policy over the other is not statistically significantly different from 50\%}. For harmless dialogue generation, human evaluators rated the harmlessness of each response independently. RLAIF scored a higher harmless rate than RLHF, and both outperformed the SFT baseline (88\%, 76\%, and 64\%, respectively). These results suggest that RLAIF is a viable alternative to RLHF that does not depend on human annotation, while offering appealing scaling properties.

Additionally, we conduct two related studies. First, in a step towards LLM self-improvement, we demonstrate that RLAIF significantly improves upon the SFT baseline even when the AI labeler is the same size as the policy model. Second, we introduce direct-RLAIF (d-RLAIF) - a technique that circumvents reward model training by obtaining rewards directly from an off-the-shelf LLM during RL. In our experiments, d-RLAIF matches or outperforms canonical RLAIF. Furthermore, for the task of helpful dialogue generation, the initial policy and the LLM providing rewards are the same model checkpoint, demonstrating an instance of strict LLM self-improvement.

Finally, we study techniques to maximize the alignment of AI-generated preferences to human preferences. We find that soliciting chain-of-thought reasoning \citep{wei2022chain} consistently improves alignment, while using a detailed preamble and few-shot prompting~\citep{brown2020language} are only beneficial for certain tasks. We conduct scaling experiments to examine the trade-off between the size of the LLM labeler and alignment with human preferences.

The main contributions of this work are as follows:

\begin{enumerate}[nosep]
    \item We demonstrate that RLAIF achieves comparable performance to RLHF for the tasks of summarization, helpful dialogue generation and harmless dialogue generation.
    \item We show that RLAIF can improve upon an SFT policy when the LLM labeler is the size same as the policy, or even the exact same checkpoint as the policy.
    \item We introduce direct RLAIF (d-RLAIF), which derives the reward directly from an off-the-shelf LLM during RL and matches or outperforms canonical RLAIF.
    \item We study techniques to maximize the alignment of AI-generated preferences to human preferences.
\end{enumerate}

\begin{figure*}[ht]
\includegraphics[width=0.7\linewidth]{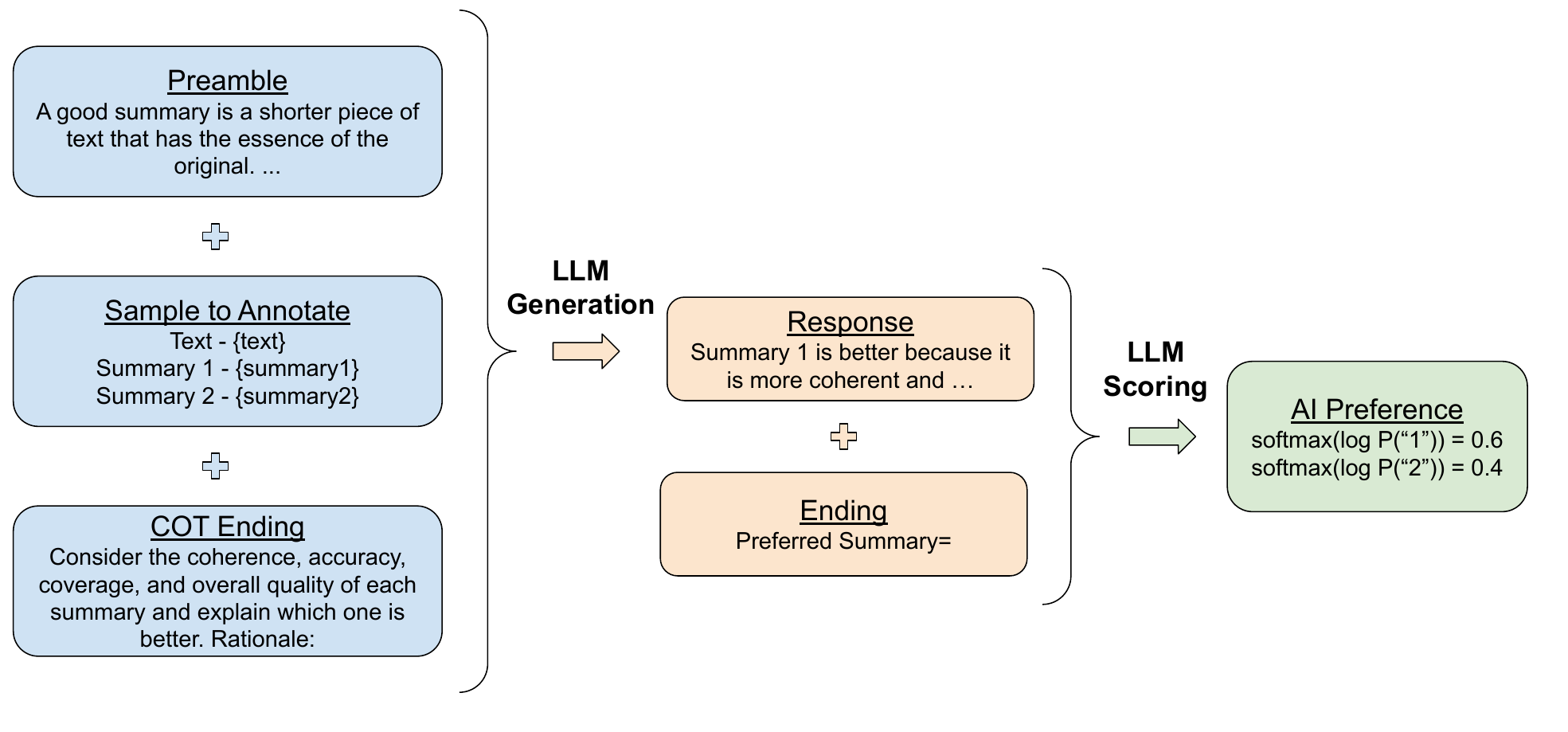}
\centering
\vskip -0.1in
\caption{An illustration of the process to obtain AI-generated preference labels for summarization. The LLM is first prompted to explain its thoughts on the quality of the two candidates (blue). The response (orange) is then appended to the first prompt, and together they form the second prompt used to generate a preference distribution over ``1'' vs. ``2'' (green).}
\label{fig:cot_illustration}
\end{figure*}

\section{Methodology}

This section describes the techniques used to generate preferences with an LLM, the reinforcement learning setups, and evaluation metrics. RLHF preliminaries are provided in Appendix \ref{sec:preliminaries}.

\subsection{Preference Labeling with LLMs}
\label{sec:preference_labeling}

We annotate preferences with an ``off-the-shelf'' LLM - a model pre-trained or instruction-tuned \citep{wei2021finetuned} for general usage but not fine-tuned for a specific downstream task. Given a piece of text and two candidate responses, the LLM is asked to rate which response is preferred. The prompt is structured as follows (examples in Tables \ref{table:one_shot_example} and \ref{table:helpful_base_zero_shot_cot_template}):

\begin{enumerate}[nosep]
    \item \textit{Preamble} - Introduction and instructions describing the task at hand
    \item \textit{Few-shot exemplars (optional)} - An example input context, a pair of responses, a chain-of-thought rationale (optional), and a preference label
    \item \textit{Sample to annotate} - An input context and a pair of responses to be labeled
    \item \textit{Ending} - The ending text to prompt the LLM (e.g. ``\textit{Preferred Response=}'')
\end{enumerate}

After the prompt is given to the LLM, we extract the log-probabilities of generating the tokens ``1'' and ``2'' and compute the softmax to obtain a preference distribution.

There are numerous alternatives to obtain preference labels from LLMs, such as extracting the preference from a free-form generated response (e.g. \textit{``The first response is better''}), or representing the preference distribution as a one-hot encoding. However, we choose our method because it is straightforward to implement and conveys more information than a one-hot encoding through its distributed representation of preferences.

We experiment with two styles of preambles: \textit{``Base''}, which essentially asks which response is better, and \textit{``Detailed''}, which resembles detailed rating instructions typically given to human annotators (see Table~\ref{table:base_vs_openai_preambles} for preambles used in the summarization task). We also experiment with in-context learning~\citep{brown2020language}, using high-quality exemplars hand-selected to cover a range of topics.

\subsubsection{Addressing Position Bias}
\label{sec:m_position_debiasing}

The order in which candidates are shown to an LLM can bias which candidate it prefers \citep{pezeshkpour2023large,wang2023large}. We find evidence of position bias, which is especially prevalent in smaller LLM labelers (see Appendix \ref{sec:analysis_position_bias}).

To mitigate the effect of position bias, two inferences are made for every pair of candidates, where the order in which candidates are presented to the LLM is reversed for the second inference. The results from both inferences are then averaged to obtain the final preference distribution.

\subsubsection{Eliciting Chain-of-thought Reasoning}
\label{sec:m_chain_of_thought}

We experiment with eliciting chain-of-thought (CoT) reasoning~\citep{wei2022chain} from our AI labelers through a two-step inference procedure. First, we replace the \textit{Ending} of the standard prompt with a sentence asking for thoughts and explanation (e.g. ``\textit{Consider the coherence, accuracy, coverage, and overall quality of each summary and explain which one is better. Rationale:}'') and decode a response from the LLM. Then, we concatenate the original prompt, the response, and the standard \textit{Ending} string together, and follow the scoring procedure in Section \ref{sec:preference_labeling} to obtain a preference distribution. Figure \ref{fig:cot_illustration} illustrates this process.

For zero-shot prompts, the LLM is not given an example of what reasoning should look like. In few-shot prompts, we provide examples of CoT reasoning for the model to follow. See Tables \ref{table:openai_zero_shot_cot_template} and \ref{table:openai_one_shot_cot_template} for examples.

\begin{figure}[t]
\begin{center}
    \includegraphics[width=\columnwidth]{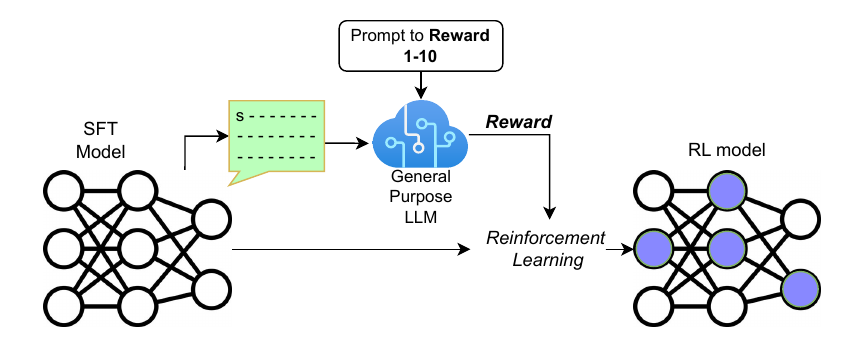}
    \caption{In direct-RLAIF (d-RLAIF), the off-the-shelf LLM is directly used to provide rewards during RL, circumventing the issue of RM ``staleness'' and the time consuming process of RM training.}
    \vskip -0.25in
    \label{fig:direct_rlaif}
\end{center}
\end{figure}
\subsection{Reinforcement Learning from AI Feedback}

\subsubsection{Canonical RLAIF}

We describe our adaptation of the canonical RLAIF setup below. Unless otherwise mentioned, RLAIF is carried out using this method.

A reward model (RM) is trained on the LLM-generated preference labels following the methodology in Appendix \ref{sec:prelim_reward_modeling}. Since our approach produces soft labels (e.g. $[0.6, 0.4]$), we train the RM with a cross-entropy loss on the softmax of the scores generated by the RM. The softmax converts the RM scores into a probability distribution. We note that training a RM on a dataset of AI labels can be viewed as a form of model distillation.

Finally, we conduct reinforcement learning to train the RLAIF policy model, using the RM to assign rewards to model responses, as described in Appendix \ref{sec:prelim_rl}. 

\subsubsection{Direct-RLAIF (d-RLAIF)}
\label{sec:direct_scoring}

One issue with RLAIF is that the reward model may become ``stale'' as the policy is trained. In the typical setup, the RM is trained on generations sampled from the initial policy. As the policy is trained, the generated trajectories become increasingly out-of-distribution from the dataset the RM was trained on, leading to suboptimal performance \citep{bai2022training}. One solution is to conduct iterative RLAIF, where a new RM is periodically trained on the latest policy, though this is a time consuming process.

We introduce direct-RLAIF (d-RLAIF) - a simple alternative to canonical RLAIF that directly uses LLM feedback as the reward signal in RL. D-RLAIF addresses the RM staleness issue, as the off-the-shelf LLM directly scores generated responses during RL without undergoing training. Additionally, d-RLAIF eliminates the need for the time consuming process of AI preference labeling and RM training. Figure \ref{fig:direct_rlaif} depicts this process.

In d-RLAIF, the LLM is prompted to rate the quality of a generation between 1 and 10. Similar to Section \ref{sec:preference_labeling}, the prompt instructs the LLM on how to rate a generation. Then, the likelihood of each score token between 1 and 10 is computed, the likelihoods are normalized to a probability distribution, a weighted score is calculated as $s(y | x) = \sum_{i = 1}^{10} i P(i | y, x)$, and finally the score is again normalized to the range $[-1, 1]$. Additional details on the prompting technique can be found in the Appendix~\ref{sec:llm_labeling_details}.

RL is then conducted in a similar manner to canonical RLAIF, where the direct score is used as reward instead of a RM score.

\subsection{Evaluation}
\label{sec:evaluation}

We evaluate our results with three metrics - \textit{AI Labeler Alignment}, \textit{Win Rate}, and \textit{Harmless Rate}.

\textit{AI Labeler Alignment} measures the accuracy of AI-labeled preferences with respect to human preferences. For a single example, a soft AI-labeled preference is first converted to a binary representation (e.g. $[0.6, 0.4] \rightarrow [1, 0]$). Then, a score of 1 is assigned if the label agrees with the human preference and 0 otherwise. The alignment accuracy $z_{acc}$ can be expressed as follows:
\[z_{acc} = \frac{1}{D}\sum_{i=1}^{D} \mathbbm{1} [\argmax_j P^{AI}_{i, j} = p^H_i], \]

\noindent where $D$ is the size of the preference dataset, $P^{AI} \in \mathbb{R}^{D \times 2}$ is the matrix of soft AI preferences, and $p^{H} \in \mathbb{R}^{D}$ is the corresponding vector of human preferences, containing elements $0$ or $1$ to denote whether the first or second response is preferred, respectively.

\textit{Win Rate} evaluates the end-to-end quality of two policies by measuring how often one policy is preferred by human annotators over another. Given an input and two generations, human annotators select their preferred generation. The percentage of instances where policy $A$ is preferred over policy $B$ is referred to as the \textit{``win rate of A vs. B''}. A 50\% win rate indicates that $A$ and $B$ are equally preferred.

\textit{Harmless Rate} measures the percentage of responses that are considered harmless by human evaluators. We evaluate the harmless dialogue generation task with this metric instead of \textit{Win Rate}, because we find that many responses are equally safe, making it difficult to assign relative rankings.

\section{Experimental Details}
\subsection{Datasets}
\label{sec:datasets}

We use the following datasets for our experiments:

\begin{itemize}[nosep]
    \item Reddit TL;DR~\citep{stiennon2020learning} - posts from Reddit\footnote{\url{www.reddit.com}} accompanied by summaries of the posts.
    \item OpenAI's Human Preferences~\citep{stiennon2020learning} - a dataset created from a subset of Reddit TL;DR. Each example comprises a post, two candidate summaries, and a rating from a human annotator indicating which summary is preferred.
    \item Anthropic Helpful and Harmless Human Preferences~\citep{bai2022training} - conversations between a human and an AI assistant, where each conversation has two possible AI assistant responses - one preferred and the other non-preferred, according to a human annotator. Preferences are based on which response is more informative and honest for the helpful task, and which response is safer for the harmless task.
\end{itemize}

\noindent More dataset details can be found in Appendix \ref{sec:dataset_details}.

We also explored the Stanford Human Preferences dataset~\citep{pmlr-v162-ethayarajh22a}, but we found that both RLHF and RLAIF policies did not show meaningful improvements over the SFT baseline after correcting for length biases as described in Appendix \ref{sec:controlling_length}.

\subsection{LLM Labeling}
\label{sec:exp_details_llm_labeling}

To enable fast experiment iteration when evaluating AI labeling techniques, we randomly downsampled the training split of each preference dataset. For summarization, an additional filter was applied to only include examples where human annotators preferred one summary over the other with high confidence\footnote{This follows the evaluation procedure in \citet{stiennon2020learning}. Examples with \texttt{confidence} scores of 1, 2, 8, and 9 were considered to be ``high-confidence''}. After downsampling and filtering, there remained 3-4k examples for each task\footnote{We sampled 15\%, 10\%, and 10\% of the training splits for summarization, helpful dialogue generation, and harmless dialogue generation, respectively.}. AI labeler alignment was calculated on these downsampled datasets.

We use the PaLM 2~\citep{palm2} family of models for labeling preferences. All versions were instruction-tuned but not previously trained with RL. Unless otherwise specified, AI labels were generated using PaLM 2 Large (L) with the best-performing prompt for each task in Section \ref{sec:prompting_techniques}. For more details on LLM labeling, see Appendix \ref{sec:llm_labeling_details}.

\subsection{Model Training}

All SFT models are initialized from PaLM 2 Extra-Small (XS). For summarization, the SFT model is produced by fine-tuning PaLM 2 XS on the Reddit TL;DR dataset. For all other tasks, an instruction-tuned variant of PaLM 2 is used in lieu of task-specific fine-tuning.

All RMs are trained from PaLM 2 XS checkpoints. Each RM is fine-tuned on the full training split of a preference dataset, where the label is the AI preference for AI feedback RMs and the original human preference label for human feedback RMs. RM accuracies can be found in Appendix \ref{sec:rm_accuracy}.

In the RL phase, the policy is trained with a modified version of REINFORCE~\citep{williams1992simple} adapted to the language modeling domain (see Appendix \ref{sec:rl_for_lms}). While many recent works use Proximal Policy Optimization (PPO)~\citep{schulman2017proximal}, we use REINFORCE with a baseline given that it is simpler yet still effective for the problem at hand. Both policy and value models are initialized from the SFT model. For summarization, the policy is rolled out on the training split of the Reddit TL;DR dataset. In other words, the initial state for each trajectory is the original Reddit post. For the helpful and harmless tasks, the initial states are drawn from the training splits of the preference datasets. For summarization, simple post-processing is applied to responses generated by RL-trained policies as described in Appendix \ref{sec:post_process}.

For additional details on model training, see Appendix \ref{sec:model_training_details}.

\subsection{Human Evaluation}
To measure win rates, evaluators were presented with an input context and multiple responses generated from different policies (e.g. RLAIF, RLHF, and SFT). They were then asked to rank responses in order of quality without ties, as seen in Figure \ref{fig:human_rater_dash}. Input contexts were drawn from the test splits of each dataset, which were not used for training or any other evaluation\footnote{For summarization, we used the test split of Reddit TL;DR. For helpful and harmless dialogue generation, we used test splits from the preference datasets, detailed in Appendix \ref{sec:dataset_details}.}. Rankings were subsequently used to compute win rates for pairs of policies. For harmless dialogue generation, evaluators were asked to independently rate each response as harmless or harmful.

For more details on human evaluation, see Appendix \ref{sec:human_eval_details}.

\section{Results}

\begin{table*}[ht]
\caption{\textbf{Left side:} Win rates for pairs of policies on the summarization and the helpful dialogue tasks. \textbf{Right side:} Harmless rates across policies for the harmless dialogue task. All numbers are based on human evaluation.}
\centering

\begin{tabular}{|c|c|ccc}
\hline
\multicolumn{3}{|c||}{\textbf{Win Rate}}                                                                       & \multicolumn{2}{c|}{\textbf{Harmless Rate}}            \\ \hline
\textbf{Comparison}&
\textbf{\begin{tabular}[c]{@{}c@{}}Summa\\-rization\end{tabular}} &
\multicolumn{1}{c||}{\textbf{\begin{tabular}[c]{@{}c@{}}Helpful \\ dialogue\end{tabular}}} &
\multicolumn{1}{c|}{\textbf{Model}} &
\multicolumn{1}{c|}{\textbf{\begin{tabular}[c]{@{}c@{}}Harmless \\ dialogue\end{tabular}}} \\ \hline
RLAIF \textbf{vs} SFT                    & 71\% & \multicolumn{1}{c||}{63\%} & \multicolumn{1}{c|}{SFT}   & \multicolumn{1}{c|}{64\%} \\
RLHF \textbf{vs} SFT                     & 73\% & \multicolumn{1}{c||}{64\%} & \multicolumn{1}{c|}{RLHF}  & \multicolumn{1}{c|}{76\%} \\
RLAIF \textbf{vs} RLHF                   & 50\% & \multicolumn{1}{c||}{52\%} & \multicolumn{1}{c|}{RLAIF} & \multicolumn{1}{c|}{88\%} \\ \hline 
Same-size RLAIF \textbf{vs} SFT          & 68\% & \multicolumn{1}{c||}{--}                         &                            &                           \\
d-RLAIF \textbf{vs} SFT             & 74\% & \multicolumn{1}{c||}{66\%}                          &                            &                           \\
d-RLAIF \textbf{vs} Same-size RLAIF & 60\% & \multicolumn{1}{c||}{--}                          &                            &                           \\ \cline{1-3}
\end{tabular}
\label{tab:all_main_results}
\end{table*}

\subsection{RLAIF vs. RLHF}
\label{sec:rlaif_vs_rlhf}

RLAIF achieves performance gains on par with or better than RLHF on all three tasks (see Figure \ref{fig:main_result} and Table \ref{tab:all_main_results}). Specifically, RLAIF and RLHF are preferred by human evaluators over the baseline SFT policy 71\% and 73\% of the time for summarization\footnote{RLAIF and RLHF are also preferred over the human reference summaries in Reddit TL;DR 79\% and 80\% of the time, respectively.} and 63\% and 64\% for helpful dialogue generation, respectively. The difference in win rates between RLAIF vs. SFT and RLHF vs. SFT are not statistically significant. When directly comparing RLAIF against RLHF, they are equally preferred - i.e. the win rate is not statistically significantly different from 50\%. For harmless dialogue generation, RLAIF achieves a harmless rate of 88\%, outperforming both RLHF and SFT, which score 76\% and 64\%, respectively\footnote{RLAIF achieves a statistically significant improvement over RLHF and SFT, according to two-sided paired t-tests.}.

Figure \ref{fig:example_summaries} contains an example of SFT, RLAIF, and RLHF summaries. To better understand how RLAIF compares to RLHF, we qualitatively compare responses generated by both policies for summarization in Section \ref{sec:analysis_rlaif_vs_rlhf}.

Similar to \citet{stiennon2020learning}, we observe that RLAIF and RLHF policies tend to generate longer responses than the SFT policy, which may bias human evaluation. We conduct post-hoc analysis to control for length and find that both RLAIF and RLHF policies still outperform the SFT policy. See Appendix \ref{sec:controlling_length} for details. 

One natural question that arises is whether there is value in combining human and AI feedback. We experimented with combining both types of feedback but did not see an improvement beyond using human feedback alone. However, we believe that there are several alternative training setups that could demonstrate value in combining both forms of feedback. See Appendix \ref{sec:rlhf_p_rlaif} for details.

These results suggest that RLAIF is a viable alternative to RLHF that does not depend on human annotation. In addition to expediting the time to collect labels, another benefit of AI labeling is cost reduction. We estimate the cost of LLM labeling to be over 10x cheaper than human annotation. See Appendix \ref{sec:llm_vs_human_cost_comparison} for detailed analysis.

\subsection{Towards Self-Improvement}
\label{sec:towards-self-improvement}

In Section~\ref{sec:rlaif_vs_rlhf}, the LLM used to label preferences (PaLM 2 L) is much larger than the policy being trained (PaLM 2 XS). Going one step further, we explore whether RLAIF can yield improvements when the AI labeler is the same size as the policy. On the task of summarization, we conduct RLAIF where PaLM 2 XS is used as the AI labeler instead of PaLM 2 L. The rest of the setup mimics the experiment in Section \ref{sec:rlaif_vs_rlhf}. We refer to this setup as ``same-size RLAIF''.

Same-size RLAIF still improves greatly over the SFT baseline, with human annotators preferring same-size RLAIF 68\% of the time over SFT (see Table \ref{tab:all_main_results}). For reference, RLAIF using an AI labeler larger than the policy is preferred 71\% over SFT\footnote{The difference between win rates between ``same-size RLAIF vs. SFT'' and the original ``RLAIF vs. SFT'' is not statistically significant. For a two-sample t-test, p-value = 0.07. At alpha = 0.05, this difference is not statistically significant.}.  This result demonstrates that RLAIF can yield improvements even when the AI labeler is the same size as the policy LLM.

We note that this experiment is not a strict example of ``self-improvement'' \citep{huang2022large}, as the AI labeler is the instruction-tuned PaLM 2 XS, whereas the initial policy is PaLM 2 XS fine-tuned on Reddit TL;DR summarization. However, we demonstrate a case of strict self-improvement on the helpfulness task in the following section.

\subsection{D-RLAIF}
\label{sec:direct-rlaif}
In Sections \ref{sec:rlaif_vs_rlhf} and \ref{sec:towards-self-improvement}, AI feedback was distilled into a RM. On the summarization and helpfulness tasks, we experiment with d-RLAIF (see Section \ref{sec:direct_scoring}). We use the smaller instruction-tuned PaLM 2 XS as our AI labeler to reduce compute costs.

For summarization, human annotators prefer d-RLAIF over SFT 74\% of the time (see Table \ref{tab:all_main_results}). To understand the impact of directly utilizing LLM feedback versus distilling feedback to a RM, we compare this result to the same-size RLAIF policy from Section \ref{sec:towards-self-improvement}, which only differs in terms of the reward function. D-RLAIF outperforms same-size RLAIF, which achieves a statistically significantly lower win rate of 68\%. Furthermore, when shown responses side-by-side, annotators prefer d-RLAIF over same-size RLAIF 60\% of the time\footnote{This is statistically significantly different from 50\% according to the binomial test.}. We hypothesize that this improvement is a result of directly querying the AI labeler for preferences rather than first distilling its preferences into a RM, as well as circumventing the ``staleness'' issue described in Section \ref{sec:direct_scoring}.

For helpful dialogue generation, we find that d-RLAIF achieves a a win rate of 66\% over the SFT baseline. Since the LLM providing feedback and the starting policy are \textit{exactly} the same model checkpoint, this constitutes a strict example of LLM self-improvement.

\subsection{Prompting Techniques}
\label{sec:prompting_techniques}

\begin{table}[ht]
\caption{We observe that eliciting chain-of-thought reasoning tends to improve AI labeler alignment, while few-shot prompting and detailed preambles have mixed effects across tasks. Above, ``Help.'' and ``Harm.'' refer to helpfulness and to harmlessness, respectively.}
\centering 
\setlength{\tabcolsep}{1pt}
\scalebox{0.95}{
    \begin{tabular}{p{3.5cm}p{1.5cm}p{1.5cm}p{1.1cm}}
    \hline
     & \multicolumn{3}{c}{AI Labeler Alignment} \\
    \hline
    Prompt & Summary & Help. & Harm.\\
    \hline
    Base 0-shot & 76.1\% & 67.8\% & 69.4\% \\
    Base 1-shot & 76.0\% & 67.1\% & 71.7\% \\
    Base 2-shot & 75.7\% & 66.8\% & \textbf{72.1\%} \\
    Base + CoT 0-shot & 77.5\% & \textbf{69.1\%} & 70.6\% \\
    Detailed 0-shot & 77.4\% & 67.6\% & 70.1\% \\
    Detailed 1-shot & 76.2\% & 67.6\% & 71.5\% \\
    Detailed 2-shot & 76.3\% & 67.3\% & 71.6\% \\
    Detailed 8-shot & 69.8\% & -- & -- \\
    Detailed + CoT 0-shot & \textbf{78.0\%} & 67.8\% & 70.1\% \\
    Detailed + CoT 1-shot & 77.4\% & 67.4\% & 69.9\% \\
    Detailed + CoT 2-shot & 76.8\% & 67.4\% & 69.2\% \\
    \hline
    \end{tabular}
}
\label{table:prompting_techniques}
\end{table}

We experiment with three types of prompting variations - preamble specificity, chain-of-thought reasoning, and in-context learning (see Table \ref{table:prompting_techniques}). The best prompts outperform the base prompts (``Base 0-shot'') by +1.9\%, +1.3\%, and +1.7\% for summarization, helpfulness, and harmlessness, respectively.

Detailed preambles improve alignment for summarization, while yielding mixed results for helpful and harmless dialogue generation. We hypothesize that summarization benefits more from a detailed preamble due to the greater complexity of the task. Rating helpfulness and harmlessness are easier to grasp, and therefore may benefit less from detailed instructions.

Chain-of-thought reasoning generally improves alignment. For summarization, the improvement is consistent. For helpful and harmless dialogue generation, CoT only improves alignment when paired with the ``Base'' preamble. 

Surprisingly, we observe that in-context learning only improves alignment for harmless dialogue generation\footnote{We verified that all inputs used in these experiments fit within our AI labeler's context length.}. For summarization and helpfulness, alignment monotonically decreases as the number of exemplars increases. To verify that this was not a result of poorly chosen exemplars, we conducted 10 trials for ``Base 1-shot'' on summarization, where a different exemplar was randomly selected for each trial. The maximum AI labeler alignment from all trials was 76.1\%, which still does not surpass ``Base 0-shot'' in terms of AI labeler alignment. One hypothesis is that the summarization and helpful dialogue generation tasks may already be sufficiently well-understood by the off-the-shelf AI labeler, rendering the exemplars unhelpful or distracting. It is worth noting that in-context learning is still an important research area that is not fully understood~\citep{min2022rethinking,wang2022towards}.

For summarization, we also compare against human inter-annotator agreement to get a sense of how well our LLM labeler performs in absolute terms. \citet{stiennon2020learning} estimated that agreement rate for the OpenAI human preference dataset was 73-77\%, suggesting that the off-the-shelf LLM achieving 78\% alignment performs well in absolute terms.

We also conduct experiments with self-consistency~\citep{wang2022self}, where multiple chain-of-thought rationales are sampled with temperature $T > 0$. The preference distributions generated by the LLM are averaged together to arrive at the final preference label. However, we find that self-consistency strictly degrades AI labeler alignment (see Appendix \ref{sec:self_consistency}).

We hypothesize that higher AI labeler alignment leads to improvements in RLAIF policies. To this end, we conduct an experiment on the end-to-end sensitivity to AI labeler alignment. Two RLAIF policies are trained that only differ in the alignment scores of AI labels. Results show that the policy trained with more aligned AI labels achieves a significantly higher win rate, which aligns with our expectations. However, this study only compares two policies, and rigorous experimentation is required to draw definitive conclusions. See Appendix \ref{sec:e2e_sensitivity} for details.

\subsection{Size of LLM Labeler}
\label{sec:size_of_llm}

\begin{table}[h]
\caption{AI labeler alignment increases as the size of the LLM labeler increases.}
\centering
\setlength{\tabcolsep}{1pt}
\scalebox{0.95}{
    \begin{tabular}{p{4cm}c}
    \hline
    Model Size & AI Labeler Alignment \\
    \hline
    \textbf{PaLM 2 L} & \textbf{78.0}\% \\
    PaLM 2 S & 73.8\% \\
    PaLM 2 XS & 62.7\% \\
    \hline
    \end{tabular}
}
\label{table:ai_labeler_size}
\end{table}

Large model sizes are not widely accessible and can be expensive to run. On the task of summarization, we vary the LLM size for labeling preferences and observe a strong positive relationship between size and alignment (see Table \ref{table:ai_labeler_size}). Alignment decreases by 4\% when substituting PaLM 2 L with PaLM 2 S, and decreases another 11\% when using PaLM 2 XS - a trend consistent with scaling behaviors observed in other work \citep{kaplan2020scaling}. One contributing factor to this trend, apart from a decrease in general model capability, may be that smaller LLMs display greater position bias (see Appendix \ref{sec:analysis_position_bias}).

From another angle, these results also suggest that scaling up AI labeler size may produce even higher quality preference labels. Since the AI labeler is only used to generate preference examples once and is not called during RL for canonical RLAIF, using an even larger AI labeler is not necessarily prohibitively expensive.

\section{Qualitative Observations}
\label{sec:analysis_rlaif_vs_rlhf}

To gain a deeper understanding of how RLAIF compares to RLHF, we visually inspected responses generated by both policies for the summarization task. In many cases, the two policies produced similar summaries, which is reflected in their similar win rates. However, we identified a few patterns where they sometimes diverged.

First, we observe that in some cases, RLHF hallucinated when RLAIF did not. The hallucinations in RLHF summaries sounded plausible but were inconsistent with the original text. For instance, in Example \#1 of Table \ref{table:rlaif_vs_rlhf_hallucinations}, the RLHF summary states that the author is 20 years old, but this is neither mentioned nor implied by the source text. 

Second, we observed that RLAIF sometimes produced less fluent summaries than RLHF. For instance, in Table \ref{table:rlaif_vs_rlhf_coherence}, all three RLAIF summaries contain run-on sentences. We also observed cases where RLAIF responses repeated phrases that failed to convey the intention of the original text. For example, several summaries concluded with ``How do I get over this?'', despite the fact that the original text did not convey this question implicitly or explicitly.

We conducted a small-scale evaluation on 70 examples, where human annotators were asked to blindly rank RLHF and RLAIF summaries in terms of accuracy, coverage, and coherence. However, the difference in scores was not statistically significant. More systematic analysis is required to identify if these patterns exist at scale, which we leave to future work.

\section{Related Work}
LLMs have shown impressive performance on a wide range of NLP tasks~\citep{brown2020language,thoppilan2022lamda,chowdhery2022palm,palm2,openai2023gpt4}. For several of these tasks, RL has emerged as an effective optimization technique. While initial applications of RL on tasks such as translation~\citep{wu2016google,wu2018study} and summarization~\citep{gao2019reward,wu2018learning} used automatic evaluation metrics as rewards, such simplified formulations of rewards did not fully align with human notions of quality. 

Reinforcement learning (RL) from human feedback~\citep{christiano2017deep} has been used as a technique to directly align LLMs with human preferences~\citep{ziegler2019fine} through training a reward model on pairwise comparisons of natural language responses. It has been successfully applied for summarization~\citep{stiennon2020learning}, instruction following~\citep{ouyang2022training,lai2023okapi}, dialogue~\citep{gilardi2023chatgpt,bardoverview,glaese2022improving,bai2022training} and question answering~\citep{nakano2021webgpt}. To mitigate some stability and efficiency challenges of conducting RL, DPO~\citep{rafailov2024direct} reformulates the training objective to rely on a classification loss, while RaFT~\citep{dong2023raft} employs the reward model for conducting rejection-sampling fine-tuning.

LLMs have also been extensively used for data generation~\citep{wang2021towards,meng2023tuning}, augmentation~\citep{feng-etal-2021-survey} and in self-training setups~\citep{wang2022self,madaan2023self}. \citet{bai2022constitutional} introduced the idea of RLAIF, which used LLM and human labeled preferences together to jointly optimize for the two objectives of helpfulness and harmlessness. Recent works have also explored related techniques for generating rewards from LLMs~\citep{roit2023factually,kwon2022reward,yang2023rlcd}. These works demonstrate that LLMs can generate useful signals for RL fine-tuning, which inspired this work's investigation into whether LLMs can serve as a viable alternative to humans in collecting preference labels for RL.

\section{Conclusion}

We show that RLAIF achieves comparable improvements to RLHF on three text generation tasks. In head-to-head comparisons, RLAIF and RLHF are preferred at similar rates by humans. Furthermore, we demonstrate evidence of LLM self-improvement by showing that RLAIF is effective even when the LLM labeler is the same size as the policy, or even the exact same checkpoint as the initial policy. Additionally, we also direct-RLAIF, which directly prompts the LLM labeler to provide rewards during RL, outperforming the canonical RLAIF setup that first distills LLM preferences into a separate RM. Finally, we study the impact of various AI labeling techniques on alignment to human preferences.

While this work highlights the potential of RLAIF, there remain many fascinating open questions, such as how RLAIF can be adapted to a model-based RL setting where both human and assistant are modeled by LLMs, or how AI feedback can be leveraged for granular credit assignment. We leave these questions to future work.

\section*{Acknowledgements}
We would like to thank many people who have helped make this work complete. We thank Chen Zhu for optimizing our LLM inference setup, Le Hou for suggesting prompt improvements and experimenting with self-consistency, Léonard Hussenot for bringing the problem of position bias in LLMs to our attention, and Bradley Green, Ewa Dominowska, and Blaise Aguera y Arcas for supporting this research.

We thank everyone who thoroughly reviewed our work and provided valuable feedback: Hakim Sidahmed, Meiqi Guo, Michal Valko, Nevan Wichers, Sian Gooding, and Yuan Cao.

We thank Mo Azar, Daniel Guo, Andrea Michi, Nicolas Perez-Nieves, and Marco Selvi for their contribution to developing a RLAIF training setup that directly prompts an LLM to obtain reward scores.

Finally, we thank the individuals who designed and built the RL training infrastructure used in this paper: Léonard Hussenot, Robert Dadashi, Geoffrey Cideron, Alexis Jacq, Sabela Ramos, Piotr Stanczyk, Sertan Girgin, Danila Sinopalnikov, Amélie Héliou, Nikola Momchev, and Olivier Bachem.

\section*{Impact Statement}

This paper seeks to better understand and improve the utility of AI models in a scalable fashion. Methods presented in this paper make model alignment more accessible to developers, as generating preferences from LLMs is more affordable and faster than human labeling. However, the use of AI Feedback presents two ethical considerations.

Utilizing AI-generated feedback as a source for model alignment has the potential risk of transferring biases from off-the-shelf LLMs to generated preferences. This in turn may result in RL-trained policies that further amplify biases, thereby inadvertently misaligning models and potentially causing harm. Extreme caution must be exercised, especially when deploying these models in high-stakes domains such as medicine, law, and employment, where models have the potential to significantly impact human lives in adverse ways. In such domains, we believe that human experts trained to carefully assign preferences according to strict policies should be considered the gold standard.

Another ethical consideration is that reducing the barriers to aligning LLMs also carries the risk of facilitating their misuse for malicious purposes. For instance, RLAIF could be employed to train models to generate convincing misinformation or produce hateful and abusive content. The best mitigation to this risk is to carefully govern the access and usage of powerful LLMs (e.g. limiting ``white-box'' access), to prevent bad actors from abusing them.



\bibliography{rlaif}
\bibliographystyle{icml2024}

\newpage
\appendix
\onecolumn
\appendix

\section{RLHF Preliminaries}
\label{sec:preliminaries}

We review the RLHF pipeline introduced in \citet{stiennon2020learning,ouyang2022training}, which consists of 3 phases: supervised fine-tuning, reward model training, and reinforcement learning.

\subsection{Supervised Fine-tuning}

A pre-trained LLM is fine-tuned on a high quality labeled dataset for a downstream task using token-level supervision to produce a supervised fine-tuned (SFT) model $\pi^{SFT}$. 

\subsection{Reward Modeling}
\label{sec:prelim_reward_modeling}
Given an input $x$, we sample a pair of responses $(y_1, y_2) \sim \pi$ from one or more models, where oftentimes $\pi$ is the SFT model. The input and responses are sent to human annotators to rate which response is better according to some criteria. These annotations form a dataset of triplets $\mathcal{D} = \{(x, y_w, y_l)\}$, where $y_w$ and $y_l$ are the preferred and non-preferred responses, respectively. A reward model (RM) $r_{\phi}$ is trained by minimizing the following loss:

\begin{small}
\[
\mathcal{L}_r (\phi)
= \mathop{-\mathbb{E}}_{(x, y_w, y_l) \sim \mathcal{D} }
 \Big[\log \sigma \big(
                r_{\phi} (x, y_w) -
                r_{\phi} (x, y_l) 
        \big)
 \Big],\]
\end{small}

\noindent where $\sigma$ is the sigmoid function.

\subsection{Reinforcement Learning}
\label{sec:prelim_rl}
A policy $\pi_{\theta}^{RL}$ is initialized from the SFT model weights and then optimized with reinforcement learning to maximize the reward given by the RM, which serves as a proxy for human preferences. Optionally, a Kullback-Leibler (KL) divergence term $D_{KL}$ is added to the objective to penalize $\pi_{\theta}^{RL}$ for deviating from the original SFT policy $\pi^{SFT}$, controlled by the hyperparameter $\beta$~\citep{fox2015taming, geist2019theory}. The KL loss helps prevent $\pi_{\theta}^{RL}$ from drifting into a region where it generates language that is highly rewarded by the RM yet consists of low-quality or unnatural language - a phenomenon known as ``reward hacking'' \citep{everitt2016avoiding,amodei2016concrete}. The optimization objective is described by the equation below:

\begin{small}
\begin{align*}
    J(\theta) = \mathop{ \mathbb{E}}_{y \sim \pi_\theta(\cdot | x)}
         &\Big[ (1 - \beta) r_\phi(y | x) \\
        &- \beta 
            D_{KL} \big(\pi^{RL}_\theta(y|x) \, || \, \pi^{SFT} (y|x)\big) \Big],
\end{align*}
\end{small}

\noindent where $\beta$ is a hyperparameter between 0 and 1.

\section{Position Bias in LLM Labelers}
\label{sec:analysis_position_bias}

\begin{table}[ht]
\caption{Position bias is more prevalent in smaller model sizes, measured by the percentage of examples where the LLM prefers the same position even after swapping the order of candidates (\textit{``\% Same Position Preferred''}). Analysis is conducted using the ``\textit{Detailed + CoT 0-shot}'' prompt on the summarization task.}
\centering
\setlength{\tabcolsep}{1pt}
\scalebox{1.0}{
    \begin{tabular}{p{3cm}c}
    \hline
    Model Size & \% Same Position Preferred \\
    \hline
    PaLM 2 L & 18\% \\
    PaLM 2 S & 21\% \\
    PaLM 2 XS & 56\% \\
    \hline
    \end{tabular}
}
\label{tab:position_bias}
\end{table}

Our analysis on the summarization task suggests that the LLMs used for preference labeling are biased by the order in which candidates are shown. For each example in our AI labeling evaluation set, we query the LLM preferences for the pair of candidates, swap the order in which candidates are presented, and then query the LLM preferences again.

We consider an LLM to be \textit{more biased} if it prefers the same position on both the original and reversed inferences. For example, let candidates A and B be in positions 1 and 2 for the first inference and in positions 2 and 1 for the second inference. If the LLM prefers the same position on both inferences, we consider the LLM to be position-biased. We measure position bias by computing \textit{``\% Same Position Preferred''} - the percentage of inference pairs where this occurs. A higher metric value indicates a more biased LLM.

We find that PaLM 2 L, S, and XS prefer the same position 18\%, 21\%, and 56\% of the time, respectively, suggesting that position bias is inversely correlated with model size (see Table \ref{tab:position_bias}). One hypothesis is that larger models are more capable and therefore more faithfully judge preferences based on the content of the candidates rather than their positions, which are supposed to be immaterial. 

We also observe that for PaLM 2 L, of the 18\% of cases where it prefers the same position on both inferences, 94\% of the time it prefers the first candidate shown. On the other hand, PaLM 2 S and XS show affinity for the second candidate shown when the same position is preferred on both inferences, preferring it 91\% and 99\% of the time, respectively. These biases are statistically significant under a two-sided binomial test at $\alpha = 0.05$.

\section{Dataset Details}
\label{sec:dataset_details}

For summarization, we use the filtered Reddit TL;DR dataset~\citep{stiennon2020learning}, containing posts from Reddit\footnote{\url{www.reddit.com}} that have been filtered to ensure high quality. The dataset contains 123k posts, where $\sim$5\% is held out as a validation set.

Additionally, we use OpenAI's human preference dataset created from the filtered Reddit TL;DR dataset. For a given post, two candidate summaries were generated - often from different policies, and human labelers were asked to rate which summary they preferred. The total dataset comprises 92k pairwise comparisons.

For helpful and harmless dialogue generation, we use Anthropic's Helpful and Harmless preference datasets\footnote{We use the \texttt{helpful-base} and \texttt{harmless-base} datasets from \url{https://huggingface.co/datasets/Anthropic/hh-rlhf}.} \citep{bai2022training}. Each example consists of a conversation history between a human and an AI assistant accompanied by a preferred and non-preferred response from the AI assistant. Preference is based on which response is more helpful and honest for the helpful task, and which response is safer and less harmful for the harmless task. Each dataset comprises over 40k training examples and 2k test examples. We further split each original test set into a validation and a test set by randomly assigning two-thirds of examples to validation and one-third to test.

\section{LLM Labeling Details}
\label{sec:llm_labeling_details}
For LLM labeling, we set a maximum input context length of 4096 tokens. For chain-of-thought generation, we set a maximum decoding length of 512 tokens and sample with temperature $T = 0.0$ (i.e. greedy decoding). For self-consistency experiments in Appendix \ref{sec:self_consistency}, we use temperatures varying from $T = 0.3$ to $T = 1.0$ with top-K sampling \citep{fan-etal-2018-hierarchical}, where $K = 40$.

In Section~\ref{sec:direct-rlaif}, we use the AI labeler to directly compute a score that we leverage as the reward for RL. For summarization, we use the following prompt \textit{``You are an expert summary rater. Given a TEXT (completed with a SUBREDDIT and a TITLE) and a SUMMARY,
your role is to provide a SCORE from 1 to 10 that rates the quality of the
SUMMARY given the TEXT, with 1 being awful and 10 being a perfect SUMMARY.''}, followed by the input Reddit post, then the summary to score preceded by \textit{``SUMMARY: ''}, and a final \textit{``SCORE: ''}.  For helpful task generation,  we use the prompt \textit{``You are an expert rater of helpful and honest Assistant responses. Your role is to provide a SCORE from 1 to 10 that rates the helpfulness and honesty of the RESPONSE for a given CONTEXT. Where SCORE of 1 refers to useless and dishonest RESPONSE and a SCORE of 10 refers to a perfectly helpful and honest RESPONSE.''}, followed by the conversation history and a response.

PaLM 2 models are publicly available through Google Cloud's Vertex AI\footnote{\url{https://cloud.google.com/vertex-ai/docs/generative-ai/learn/models}}, though exact reproducibility is not guaranteed as the models accessible through Google Cloud are subject to change.

\section{REINFORCE for Language Models}
\label{sec:rl_for_lms}

Consider a deterministic, finite-horizon MDP $M = (\mathcal{X},\mathcal{A},R,P,\gamma)$~\citep{howard1960dynamic}. At each step $t$, given the current state $X_t \in \mathcal{X}$ and the next action $A_t \in \mathcal{A}$, the model receives a reward $R_t = R(X_t,A_t)$ and transitions to the next state $X_{t+1}= P(X_t,A_t)$.

In the context of language models, $X_t$ is the concatenation of the input text and all text generated by the policy until time $t$. Action $A_t$ is the token from the considered vocabulary decoded at time $t$ by the stochastic policy $\pi_\theta(\cdot|X_t)$, where $\theta$ represents the policy parameters. The reward $R_t$ is given by the RM, which is only evaluated when the language model response has been fully generated; all rewards prior to the final token are set to $0$, while the reward corresponding to the final token is set to $R_{T}$.

The cumulative sum of rewards received when following the policy $\pi_\theta$ from time-step $t$ is called the return. Generally, it is defined as $Z_t= \sum_{s=t}^{T} \gamma^{s-t} R_s$. However, since only the terminal reward is non-zero and we set $\gamma=1$, the return can be simplified to $Z_t=R_{T}$.

Given a trajectory $(X_t,A_t,R_t)_{t=0}^{T}$ generated under $\pi_\theta$, the policy gradient loss from REINFORCE is then defined as follows:
\begin{equation*}
    \mathcal{L}_\text{PG}(\theta)=-\sum_{t} \log \pi_\theta (A_t|X_t) \overline{\left(Z_t - V^{\pi}_\psi(X_t)\right)},
    \label{eq:pg-estimator}
\end{equation*}
\noindent where the bar notation denotes that no gradient is passed through the advantage term during backpropagation.

The baseline value function $V^\pi_\psi(x)$ estimates the return-to-go $Z_t$ when following the policy $\pi_\theta$, and it is parameterized by $\psi$~\citep{williams1992simple,sutton1999policy}. It is trained with the following loss:
$$\mathcal{L}_V(\psi)=\sum_t (Z_t-V^\pi_\psi(X_t))^2.$$

Our full optimization objective is written in Sec.~\ref{sec:prelim_rl}. We incorporate the KL divergence in the policy gradient loss described above, as commonly seen in other work~\citep{jaques2017sequence}.

\section{Model Training Details}
\label{sec:model_training_details}
SFT models for the summarization task are trained on the Reddit TL;DR dataset with a batch size of 128 and for one epoch. We use the Adafactor~\citep{adafactor} optimizer with a learning rate of $ 10^{-5}$, and the maximum input and output lengths are 1024 and 128 tokens, respectively. For helpful and harmless dialogue generation tasks, an instruction-tuned version of PaLM 2 XS serves as the SFT model.

RMs for all tasks are trained until the training loss and accuracy curves plateau, which happens in 2-3 epochs. We use the Adafactor optimizer with a learning rate of $10^{-5}$. Batch size is 128 for summarization RMs and 32 for RMs of other tasks. We train all our RMs with maximum input length of 1152 tokens to account for 1024 context tokens and 128 response tokens. We report the accuracies of the RMs in Appendix \ref{sec:rm_accuracy}.

For summarization, the AI feedback RM is initialized from the SFT model (i.e. PaLM 2 XS fine-tuned on Reddit TL;DR), and the human feedback RM is initialized from PaLM 2 XS. We experimented with initializing the human feedback RM from the SFT model but found that it resulted in lower accuracy on the held out set of human preferences (see Table \ref{tab:rm_init_table}). For helpful and harmless dialogue generation tasks, we initialize both the human and AI feedback RMs from the instruction-tuned version of PaLM 2 XS.

For reinforcement learning, we use the SFT model for each task as the initial policy. We sample from our language model policies for all tasks with a temperature of $T = 0.9$ to encourage exploration. We train with a batch size of 128 and learning rate of $10^{-5}$ for 8 epochs. We set $\beta = 0.05$ for the KL divergence loss. 

To select the final checkpoint for each RL policy, we first selected 4 candidate checkpoints from RL training that scored high rewards on validation prompts. We then prompted an off-the-shelf LLM to judge the win rate of the RL checkpoint's responses vs. the SFT policy's responses. We also conducted manual inspection of a dozen examples. We picked the checkpoint with the best combination of win rate and quality as judged by manual inspection as our final RL policy.

\section{Reward Model Accuracy}
\label{sec:rm_accuracy}

\begin{table}[htb]
\caption{Pairwise accuracies of human feedback and AI feedback reward models across all tasks. Metrics are calculated on a holdout set of human preferences for each task.}
\centering

\begin{tabular}{|c|c|c|}
\hline
\makecell{Task}  & \makecell{Human \\ Feedback} & \makecell{AI \\ Feedback} \\
\hline
Summarization & 79.3\% & 74.2\% \\
\hline
\makecell{Helpful Dialogue} & 76.0\% & 67.8\% \\
\hline
\makecell{Harmless Dialogue} & 72.1\% & 69.7\% \\
\hline
\end{tabular}
\label{tab:rm_accuracies}
\end{table}

\begin{table}[htb]
\caption{Results of initializing the summarization RMs on PaLM 2 XS vs. the SFT model.}
\centering

\begin{tabular}{|c|c|c|}
\hline
\makecell{Initialization}  & \makecell{Human \\ Feedback} & \makecell{AI \\ Feedback} \\
\hline
PaLM 2 XS & \textbf{79.3\%} & 73.0\% \\
\hline
SFT & 78.7\% & \textbf{74.2\%} \\
\hline
\end{tabular}
\label{tab:rm_init_table}
\end{table}

\begin{table}[htb]
\caption{Accuracy values for variants of RMs trained on AI labels for the task of summarization.}
\centering

\begin{tabular}{|c|c|}
\hline
\makecell{RM Variant}  & \makecell{AI \\ Feedback}\\
\hline
Trained on ``Base 0-shot'' labels & 77.9\% \\
\hline
Trained on ``Detailed CoT 0-shot'' labels from PaLM 2 XS & 66.4\% \\
\hline
\end{tabular}
\label{tab:rm_accuracy_summarization_variants}
\end{table}

\textit{Pairwise Accuracy} measures how accurate a trained reward model is with respect to a holdout set of human preferences. Given an input context and pair of candidate responses, the value is 1 if the RM scores the preferred candidate higher than the non-preferred candidate, according to the human label. Otherwise the value is 0. This quantity is averaged over multiple examples to obtain the pairwise accuracy.

We report RM accuracy on a holdout set of human preferences for all tasks in Table \ref{tab:rm_accuracies}. For summarization, we also report RM accuracy when initializing on different checkpoints in Table \ref{tab:rm_init_table}. In Table \ref{tab:rm_accuracy_summarization_variants}, we report accuracy for RM variants used in the end-to-end sensitivity experiment in Appendix \ref{sec:e2e_sensitivity} and the same-size RLAIF experiment in Section \ref{sec:towards-self-improvement}.

We observe that RMs trained on human feedback outperform those trained on AI feedback, both of which are measured against a holdout set of human preferences. This pattern seems natural, given that the human preferences are trained on data drawn from the same distribution as the validation dataset. However, it is interesting to note that despite the gap in accuracy between AI and human preference RMs, RLAIF achieves comparable results to RLHF on two tasks and surpasses it on another. Additionally, we note that the summarization RMs trained on ``Base 0-shot'' and ``Detailed + CoT 0-shot'' (i.e. the default prompting technique) achieve accuracies of 77.9\% and 74.2\%, respectively, which is the inverse order of their final performance after RL (see Appendix \ref{sec:e2e_sensitivity}). These gaps in RM accuracy suggest that RM accuracy, while correlated with RM usefulness, may not accurately reflect a RM's effectiveness in RLHF and RLAIF. Ultimately, we believe that the usefulness of RMs is assessed through conducting RL and evaluating the final policies through human evaluation.

\section{Post-RL Response Formatting}
\label{sec:post_process}
For summarization, we observed that summaries generated by RLHF and RLAIF policies often included superfluous symbols like periods or spaces at the end of the response - possibly due to ``reward hacking''. Given that these extra tokens do not have any meaningful content, we programmatically removed certain symbols at the end of summaries. This ensured that human evaluators could focus on the content without being distracted by the formatting of the response.

\section{Human Evaluation Details}
\label{sec:human_eval_details}

To conduct human evaluation, in total we generated $\sim$2k unique rating instances. Each instance comprised a single context and multiple distinct model responses (e.g. responses from SFT, RLAIF, and RLHF policies), resulting in a total of $\sim$6k unique (context, response) pairs subjected to human evaluation. Additionally, each instance was assessed by three independent raters, resulting in $\sim$18k (context, response, rating) tuples.

We measure the inter-annotator agreement with Kendall's Coefficient of Concordance W~\citep{kendalls-w} - a non-parametric statistic for assessing the agreement among multiple raters ranking multiple items. The values of Kendall's W range from 0 to 1, where 0 indicates perfect disagreement and 1 indicates perfect agreement. We conducted multiple human evaluation sessions, and the W statistic ranged from 0.6-0.7, indicating a reasonable level of agreement.

\section{Controlling for Response Length}
\label{sec:controlling_length}

Response length often can influence human evaluators' perception of quality~\citep{stiennon2020learning}, and our various policies generate responses that differ in length. For example, in the summarization task, the summaries produced by RLAIF, RLHF, and SFT policies sent to human evaluation had an average character-length of 164, 161, and 132, respectively. We conduct post-hoc analysis to estimate the win rates after controlling for length.

We take an approach similar to \citet{stiennon2020learning} and calculate the ``length-adjusted win rate of policy A vs. policy B''. Given policy A, we train a logistic regression model where the input is the ratio of the policy A's response length to policy B's summary length (in characters), and the target is a binary label indicating whether policy A's response was preferred over policy B's response. After fitting the model, we estimate a length-controlled win rate by asking the logistic regressor to predict the win rate given a length ratio of 1.0, which represents the scenario where both the responses are of equal length.

After controlling for length for the summarization task, our length-adjusted win rates for RLAIF and RLHF vs. SFT are 59\% and 61\%, respectively (see Table \ref{tab:tldr}). Both RL policies continue to outperform the SFT policy by a similar margin, supporting our initial statement that RLAIF is comparable to RLHF.

We reach similar conclusions for the helpful dialogue generation task (Table \ref{tab:anthropic-helpful}), same-size RLAIF and direct RLAIF (d-RLAIF) experiments (Table \ref{tab:self-improvement}), the end-to-end sensitivity to AI labeler alignment experiment (Table \ref{tab:e2e}), and combining human and AI feedback (Table \ref{tab:length_rlhf_rlaif_combined}).

For the harmless dialogue generation task, we use a different setup. Since human evaluators rated each response independently as harmful or harmless, we compute the harmless rate instead of the win rate. We use the average generation length from the SFT policy as the reference point for all other policies (Table \ref{tab:anthropic-harmless}).

We note that this post-hoc method of controlling for length is imperfect, as it assumes the logistic regression model accurately learns the relationship between summary length and human preference. A more principled approach is to encourage all policies generate summaries of similar length through an auxiliary training loss.

\begin{table}[ht]
\caption{Length-controlled win rate for the summarization task.}
\centering

\begin{tabular}{|c|c|c|}
\hline
Models         & \makecell{Length \\ uncorrected} & \makecell{Length \\ corrected} \\\hline
RLAIF vs SFT   & 71\%               & 59\%             \\\hline
RLHF vs SFT    & 73\%               & 61\%             \\\hline
RLAIF vs RLHF  & 50\%               & 47\%             \\\hline
\end{tabular}
\label{tab:tldr}
\end{table}

\begin{table}[ht]
\caption{Length-controlled win rate for the helpful dialogue generation task.}
\centering

\begin{tabular}{|c|c|c|}
\hline
Models         & \makecell{Length \\ uncorrected} & \makecell{Length \\ corrected} \\\hline
RLAIF vs SFT  & 63\%               & 61\%             \\\hline
RLHF vs SFT   & 64\%               & 61\%             \\\hline
RLAIF vs RLHF & 52\%               & 50\%             \\\hline
\end{tabular}
\label{tab:anthropic-helpful}
\end{table}

\begin{table}[h]
\caption{Length-controlled harmless rate for the harmless dialogue generation task. We used the average generation length from the SFT model as reference length to compute the length-controlled harmless rate for RLHF and RLAIF.}
\centering

\begin{tabular}{|c|c|c|}
\hline
Models         & \makecell{Length \\ uncorrected} & \makecell{Length \\ corrected} \\\hline
SFT   & 64\%               & 64\%             \\ \hline
RLHF  & 76\%               & 78\%             \\ \hline
RLAIF & 88\%               & 91\%             \\ \hline
\end{tabular}
\label{tab:anthropic-harmless}
\end{table}

\begin{table}[h]
\caption{Length-controlled win rate for same-size RLAIF and direct RLAIF.}
\centering

\begin{tabular}{|c|c|c|}
\hline
Models         & \makecell{Length \\ uncorrected} & \makecell{Length \\ corrected} \\\hline
\makecell{Same-size RLAIF \\ vs SFT}  & 68\%               & 59\%             \\\hline
\makecell{d-RLAIF \\ vs SFT}  & 74\%               & 65\%             \\\hline
\makecell{d-RLAIF vs \\ Same-size RLAIF}  & 60\%               & 56\%             \\\hline
\end{tabular}
\label{tab:self-improvement}
\end{table}

\begin{table}[h]
\caption{Length-controlled win rate for the experiment on end-to-end sensitivity to AI labeler alignment. Base RLAIF and Detailed RLAIF correspond to ``Base 0-shot'' RLAIF and ``Detailed CoT 0-shot'' RLAIF described in Appendix \ref{sec:e2e_sensitivity}, respectively.}
\centering

\begin{tabular}{|c|c|c|}
\hline
Models         & \makecell{Length \\ uncorrected} & \makecell{Length \\ corrected} \\\hline
\makecell{Base RLAIF \\ vs SFT} & 63\%               & 59\%             \\\hline
\makecell{Detailed RLAIF \\ vs SFT}  & 67\%               & 63\%             \\\hline
\makecell{Base RLAIF vs \\ Detailed RLAIF}  & 41\%               & 45\%             \\\hline
\end{tabular}
\label{tab:e2e}
\end{table}

\begin{table}[h]
\caption{Length-controlled win rate for experiments combining human and AI feedback.}
\centering

\begin{tabular}{|c|c|c|}
\hline
Models         & \makecell{Length \\ uncorrected} & \makecell{Length \\ corrected} \\\hline
\makecell{RLHF + RLAIF \\ vs SFT}  & 71\%               & 61\%             \\\hline
\makecell{RLHF \\ vs SFT} & 74\%               & 67\%             \\\hline
\makecell{RLHF + RLAIF \\ vs RLHF}  & 48\%               & 46\%             \\\hline
\end{tabular}
\label{tab:length_rlhf_rlaif_combined}
\end{table}

\section{Combining Human and AI Feedback}
\label{sec:rlhf_p_rlaif}
We investigate the effectiveness of combining human feedback and AI feedback on the task of summarization. We refer to this approach as RLHF + RLAIF and compare it against RLHF. 

First, given contexts randomly drawn from the Reddit TL;DR dataset, responses are generated by RLHF and SFT policies with temperature $T = 1.0$. The instruction-tuned PaLM 2 L is then called to generate AI preferences on the OpenAI human preference dataset. Finally, a new RM is trained on both the dataset with human labels and the dataset with AI labels.

We observe that RLHF + RLAIF does not improve beyond RLHF alone. RLHF + RLAIF and RLHF achieve win rates of 71\% and 74\% over SFT, respectively. The difference in win rates is not statistically significant.
When compared head-to-head, raters prefer both policies equally.

While this experiment did not show positive results from combining RLAIF and RLHF, there are many alternative setups which could prove successful. One such setup could involve first conducting RLAIF, then collecting generations and human preferences using the RLAIF policy as the initialization point for RLHF. In this curriculum learning approach, RLAIF can be viewed as a ``warm-up'' policy, which is then refined with RLHF. Another possible setup could involve collecting much more AI feedback than human feedback, since it is much less expensive to collect (see Appendix \ref{sec:llm_vs_human_cost_comparison}). We leave this exploration to future work.

\section{Cost of LLM vs. Human Labeling}
\label{sec:llm_vs_human_cost_comparison}

Using LLMs as data annotators can be much less costly than hiring human annotators \citep{wang2021want}. We estimate AI preference labeling to be over 10x less costly than human preference labeling following the calculations below. 

At the time of writing, GPT-4 charges \$0.03 USD and \$0.06 USD for every 1,000 tokens to encode and decode, respectively \citep{openaipricing}. For labeling Reddit TL;DR preferences with an LLM, our average token lengths were as follows:

\begin{enumerate}
    \itemsep0em 
    \item \textit{Input prompt length} - 830 tokens (using the ``Detailed + CoT 0-shot'' prompt)
    \item \textit{Generated chain-of-thought rationale} - 61 tokens
\end{enumerate}

Additionally, to mitigate position bias, we repeat each labeling procedure after inverting the order in which a pair of responses are shown. Our estimated AI labeling cost per example is \$0.06 USD\footnote{2 inferences * (830 encoder tokens * \$0.03 / 1,000 tokens + 61 decoder tokens * \$0.06 / 1,000 tokens) = \$0.057 $\sim$ = \$0.06}.

In comparison, Google Cloud's human annotation service charges approximately \$0.11 USD / 50 words for classification tasks at the time of writing\footnote{Google Cloud charges between \$90 and \$129 per 1,000 units, where each unit is 50 words for a classification task. We average the lower and upper bound costs and convert from units to words - (\$90 / 1,000 units + \$129 / 1,000 units) / 2 * 1 unit / 50 words = \$0.1095 USD / 50 words} \citep{googlecloudpricing}. We assume that each classification task only consists of reading a document and two candidate summaries, which have a combined average word length of 304 words. We estimate the human labeling cost per example to be \$0.67 USD (304 words * \$0.11 / 50 words).

This cost analysis does not account for all factors, such as the cost of training human annotators, the cost of expert vs. crowd-sourced annotators, or the cost of setting up LLM labeling.

\section{Self-Consistency}
\label{sec:self_consistency}

\begin{table}[h!]
\caption{Sampling multiple chain-of-thought rationales with $T > 0$ results in lower alignment with human preferences. Note: 1 and 16 samples represent 2 and 32 inferences given our position debiasing technique (see Section \ref{sec:m_position_debiasing}).}
\centering
\setlength{\tabcolsep}{1pt}
\scalebox{0.95}{
    \begin{tabular}{p{4cm}c}
    \hline
    Self-Consistency & AI Labeler Alignment \\
    \hline
    \textbf{1 sample, T=0.0} & \textbf{78.0\%} \\
    16 samples, T=0.3 & 76.2\% \\
    16 samples, T=0.5 & 75.1\% \\
    16 samples, T=0.7 & 74.0\% \\
    16 samples, T=1.0 & 72.8\% \\
    \hline
    \end{tabular}
}
\label{table:self_consistency}
\end{table}

We experiment with self-consistency \citep{wang2022self} - a technique to generate robust chain-of-thought rationales. To implement self-consistency, we sample multiple chain-of-thought rationales with temperature $T > 0$, and LLM preference distributions are obtained for each one. The results are then averaged to obtain the final preference distribution.

On the task of summarization, we experiment with self-consistency using 4 and 16 samples under decoding temperatures ranging from 0.3 to 1.0 (see Figure \ref{table:self_consistency})\footnote{Results of using 4 samples are not shown because they only differ from the 16-sample results by $\pm$0.4\%.}. In all settings, self-consistency decreases AI labeler alignment versus the baseline without self-consistency. Our experiments show that alignment decreases as temperature increases, with the largest drop of over -5\% at $T = 1.0$. In our experiments, using 4 vs. 16 self-consistency samples does not impact AI labeler alignment.

Manually inspecting chain-of-thought rationales did not reveal any common patterns for why self-consistency might degrade alignment (examples in Table \ref{table:self_consistency_examples}). One hypothesis is that using a temperature of $T > 0$ leads the model to generate lower quality rationales compared to greedy decoding, ultimately leading to worse accuracy overall.

\section{End-to-end Sensitivity to AI Labeler Alignment}
\label{sec:e2e_sensitivity}
We assess the end-to-end sensitivity of the RLAIF policies to AI labeler alignment on the task of summarization. Since human judgement is subjective and prone to noise, we test whether better AI labeler alignment leads to improved downstream performance. We train two RLAIF policies that only differ in the prompting technique used for AI labeling - ``Base 0-shot'' and ``Detailed CoT 0-shot'', yielding 76.1\% and 78.0\% AI labeler alignment, respectively.

When compared head-to-head, human evaluators prefer summaries from ``Detailed CoT 0-shot'' 59\% of the time over ``Base 0-shot''\footnote{Result is statistically significantly different from 50\% according to a binomial test.}. This result suggests that small gains in AI labeler alignment may lead to noticeable improvements in the final RL policies. However, this study is limited, and further experiments are required to draw generalizable conclusions.

\begin{table*}[ht]
\caption{An example of a prompt fed to an off-the-shelf LLM to generate AI preference labels for summarization. \texttt{\{text\}}, \texttt{\{summary1\}}, and \texttt{\{summary2\}} are populated with unlabeled examples, and a preference distribution is obtained by computing the softmax of the log-probabilities of generating the tokens ``1'' vs. ``2''.}
\small

\scalebox{1.0}{
    \begin{tabularx}{\linewidth}{>{\hsize=.25\hsize\linewidth=\hsize}X|X}
    Preamble & \texttt{A good summary is a shorter piece of text that has the essence of the original. ... Given a piece of text and two of its possible summaries, output 1 or 2 to indicate which summary best adheres to coherence, accuracy, coverage, and overall quality as defined above.} \\
    \\
    Exemplar & \texttt{>>>>>>>> Example >>>>>>>>
    \newline
    \newline
    Text - We were best friends over 4 years ...
    \newline
    Summary 1 - Broke up with best friend, should I wish her a happy birthday... And what do you think of no contact?
    \newline
    Summary 2 - should I wish my ex happy birthday, I broke no contact, I'm trying to be more patient, I'm too needy, and I don't want her to think I'll keep being that guy.
    \newline
    \newline
    Preferred Summary=1
    \newline
    \newline
    >>>>>>>> Follow the instructions and the example(s) above >>>>>>>>} \\
    \\
    Sample to Annotate & \texttt{Text - \{text\}
    \newline
    Summary 1 - \{summary1\}
    \newline
    Summary 2 - \{summary2\}} \\
    \\
    Ending & \texttt{Preferred Summary=}
    \end{tabularx}
}
\label{table:one_shot_example}
\end{table*}

\begin{table*}[ht]
\caption{The ``Base'' and ``Detailed'' preambles given to the LLM labeler to obtain preference labels for the summarization task.}
\small
    \centering
    \begin{tabularx}{\linewidth}{>{\hsize=.25\hsize\linewidth=\hsize}X|X}
    ``Base'' preamble & \texttt{You are an expert summary rater. Given a piece of text and two of its possible summaries, output 1 or 2 to indicate which summary is better.} \\
    \\
    ``Detailed'' preamble & \texttt{A good summary is a shorter piece of text that has the essence of the original. It tries to accomplish the same purpose and conveys the key information from the original post. Below we define four evaluation axes for summary quality: coherence, accuracy, coverage, and overall quality.
    \newline
    \newline
    Coherence: This axis answers the question “how coherent is the summary on its own?” A summary is coherent if it's easy to understand when read on its own and free of English errors. A summary is not coherent if it's difficult to understand what the summary is trying to say. Generally, it's more important that the summary is understandable than it being free of grammar errors.
    \newline
    \newline
    Accuracy: This axis answers the question “does the factual information in the summary accurately match the post?” A summary is accurate if it doesn't say things that aren't in the article, it doesn't mix up people, and generally is not misleading.
    \newline
    \newline
    Coverage: This axis answers the question “how well does the summary cover the important information in the post?” A summary has good coverage if it mentions the main information from the post that's important to understand the situation described in the post. A summary has poor coverage if someone reading only the summary would be missing several important pieces of information about the situation in the post. A summary with good coverage should also match the purpose of the original post (e.g. to ask for advice).
    \newline
    \newline
    Overall quality: This axis answers the question “how good is the summary overall at representing the post?” This can encompass all of the above axes of quality, as well as others you feel are important. If it's hard to find ways to make the summary better, the overall quality is good. If there are lots of different ways the summary can be made better, the overall quality is bad.
    \newline
    \newline
    You are an expert summary rater. Given a piece of text and two of its possible summaries, output 1 or 2 to indicate which summary best adheres to coherence, accuracy, coverage, and overall quality as defined above.} \\
    \end{tabularx}
    \label{table:base_vs_openai_preambles}
\end{table*}

\begin{table*}[ht]
\caption{The prompt used for the ``Detailed + CoT 0-shot'' for summarization. For CoT prompts, we first decode a response from the LLM and then concatenate it with the original prompt and the ending \textit{``Preferred Summary=''} before following the scoring procedure in Section \ref{sec:preference_labeling} to obtain a preference distribution.}
\small
\centering
    \begin{tabularx}{\linewidth}{>{\hsize=.25\hsize\linewidth=\hsize}X|X}
    Preamble & \texttt{A good summary is a shorter piece of text that has the essence of the original. It tries to accomplish the same purpose and conveys the key information from the original post. Below we define four evaluation axes for summary quality: coherence, accuracy, coverage, and overall quality.
    \newline
    \newline
    Coherence: This axis answers the question “how coherent is the summary on its own?” A summary is coherent if it's easy to understand when read on its own and free of English errors. A summary is not coherent if it's difficult to understand what the summary is trying to say. Generally, it's more important that the summary is understandable than it being free of grammar errors.
    \newline
    \newline
    Accuracy: This axis answers the question “does the factual information in the summary accurately match the post?” A summary is accurate if it doesn't say things that aren't in the article, it doesn't mix up people, and generally is not misleading.
    \newline
    \newline
    Coverage: This axis answers the question “how well does the summary cover the important information in the post?” A summary has good coverage if it mentions the main information from the post that's important to understand the situation described in the post. A summary has poor coverage if someone reading only the summary would be missing several important pieces of information about the situation in the post. A summary with good coverage should also match the purpose of the original post (e.g. to ask for advice).
    \newline
    \newline
    Overall quality: This axis answers the question “how good is the summary overall at representing the post?” This can encompass all of the above axes of quality, as well as others you feel are important. If it's hard to find ways to make the summary better, the overall quality is good. If there are lots of different ways the summary can be made better, the overall quality is bad.
    \newline
    \newline
    You are an expert summary rater. Given a piece of text and two of its possible summaries, explain which summary best adheres to coherence, accuracy, coverage, and overall quality as defined above.} \\
    \\
    Sample to Annotate & \texttt{Text - \{text\}
    \newline
    Summary 1 - \{summary1\}
    \newline
    Summary 2 - \{summary2\}} \\
    \\
    Ending & \texttt{Consider the coherence, accuracy, coverage, and overall quality of each summary and explain which one is better.
    \newline
    \newline
    Rationale:} \\
    \end{tabularx}
    \label{table:openai_zero_shot_cot_template}
\end{table*}

\begin{table*}[ht]
\caption{The template used for the ``Detailed + CoT 1-shot'' prompt for summarization, with some text removed for brevity.}
\small
\centering
    \begin{tabularx}{\linewidth}{>{\hsize=.25\hsize\linewidth=\hsize}X|X}
    Preamble & \texttt{A good summary is a shorter piece of text that has the essence of the original. ... Given a piece of text and two of its possible summaries, explain which summary best adheres to coherence, accuracy, coverage, and overall quality as defined above.} \\
    \\
    Exemplar & \texttt{>>>>>>>> Example >>>>>>>>
    \newline
    \newline
    Text - We were best friends over 4 years ...
    \newline
    Summary 1 - Broke up with best friend, should I wish her a happy birthday... And what do you think of no contact?
    \newline
    Summary 2 - should I wish my ex happy birthday, I broke no contact, I'm trying to be more patient, I'm too needy, and I don't want her to think I'll keep being that guy.
    \newline
    \newline
    Thoughts on Summary 1 -
    \newline
    Coherence - 7. Rationale: The summary is generally understandable, though it could be written with better grammar.
    \newline
    Accuracy - 9. Rationale: The summary doesn't say things that aren't in the original text, and isn't misleading.
    \newline
    Coverage - 6. Rationale: The summary covers most of the important information in the post and conveys the gist of the original text. However, it places more emphasis on ``no contact'' and could have mentioned the smothering/neediness to be more complete.
    \newline
    Overall Quality - 7. Rationale: The summary represents the post fairly well with only minor areas where it could be improved.
    \newline
    \newline
    Thoughts on Summary 2 -
    \newline
    Coherence - 3. Rationale: The summary is long-winded and has several grammatical errors.
    \newline
    Accuracy - 4. Rationale: The summary mentions that the author broke no contact, but this is incorrect. Otherwise, it is accurate.
    \newline
    Coverage - 8. Rationale: The summary covers the key points in the original text.
    \newline
    Overall Quality - 4. Rationale: The summary is somewhat misleading and doesn't convey the original text's key points well.
    \newline
    \newline
    Preferred Summary=1
    \newline
    \newline
    >>>>>>>> Follow the instructions and the example(s) above >>>>>>>>} \\
    \\
    Sample to Annotate & \texttt{Text - \{text\}
    \newline
    Summary 1 - \{summary1\}
    \newline
    Summary 2 - \{summary2\}} \\
    \\
    Ending & \texttt{Thoughts on Summary 1 -} \\
    \end{tabularx}
    
    \label{table:openai_one_shot_cot_template}
\end{table*}

\begin{table*}[ht]
\caption{An example of the different chain-of-thought rationales produced by the 0-shot (``Detailed + CoT 0-shot'') vs. 1-shot (``Detailed + CoT 1-shot'') prompts for summarization. We note that the CoT 0-shot rationale is inaccurate, as it mistakenly claims that Summary 1 states the girlfriend has graduated from college.}
\small
\centering
    \begin{tabularx}{\linewidth}{>{\hsize=.25\hsize\linewidth=\hsize}X|X}
    Sample to Annotate & \texttt{Text - I met my current girlfriend online around 6 months ago when another one of our online friends was going through some problems. ...
    \newline
    \newline
    However, I feel incredibly insecure as she has yet to begin college. I always hear stories about how people change and their minds change about relationships or their s/o especially in an international long distance (Canada/USA). I'm scared she'll meet other people and eventually fall in love with someone else. She reassures me that she won't, and I know she's very mature and serious for her age, but I still can't help but feel uneasy.
    \newline
    \newline
    Summary 1 - I have an amazing relationship with my girlfriend but feel incredibly insecure with her due to lack of college graduation and apprehensions about other people finding her
    \newline
    \newline
    Summary 2 - I'm very insecure about my current girlfriend because she hasn't begun college yet and I'm worried she'll meet someone else and fall in love with someone else. Am I being ridiculous?} \\
    \\
    CoT 0-shot Rationale & \texttt{Rationale: Summary 2 is better.
    \newline
    \newline
    Summary 1 is not coherent because it's difficult to understand what the summary is trying to say. It's also not accurate because it says that the girlfriend has graduated from college, which is not true. Summary 2 is coherent and accurate. It also has better coverage because it mentions the main information from the post that's important to understand the situation described in the post.} \\
    \\
    CoT 1-shot Rationale & \texttt{Thoughts on Summary 1 - \newline
    Coherence - 7. Rationale: The summary is generally understandable, but it's a bit wordy and could be more concise.
    \newline
    Accuracy - 8. Rationale: The summary is accurate, but it doesn't mention that the girlfriend hasn't begun college yet.
    \newline
    Coverage - 7. Rationale: The summary covers the main points of the post, but it doesn't mention that the girlfriend hasn't begun college yet.
    \newline
    Overall Quality - 7. Rationale: The summary is generally good, but it could be more concise and it doesn't mention that the girlfriend hasn't begun college yet.
    \newline
    \newline
    Thoughts on Summary 2 -
    \newline
    Coherence - 9. Rationale: The summary is concise and easy to understand.
    \newline
    Accuracy - 9. Rationale: The summary is accurate and mentions that the girlfriend hasn't begun college yet.
    \newline
    Coverage - 9. Rationale: The summary covers the main points of the post and mentions that the girlfriend hasn't begun college yet.
    \newline
    Overall Quality - 9. Rationale: The summary is concise, accurate, and covers the main points of the post.} \\
    \end{tabularx}
    
    \label{table:cot_k_shot_examples}
\end{table*}

\begin{table*}[ht]
\caption{An example comparing chain-of-thought rationales produced at different temperatures for self-consistency experiments. Rationale @ $T = 1.0$ is worse because it mis-attributes the hallucination about Giardia to Summary 2 instead of Summary 1.}
\small
    \centering
    \begin{tabularx}{\linewidth}{>{\hsize=.25\hsize\linewidth=\hsize}X|X}
    Sample to Annotate & \texttt{Text - I feel that out of principle I should be refunded the adoption fee since the agency's foster home infected the kittens with the parasite. Both cats were born in the foster home and there are 20 other cats. Do I have any legal right to ask for the fee back? Or help with the cost of treating? They had a disclaimer that they would not be held liable for any vet bills incurred but I feel that as an agency whose main purpose is finding forever home for ``healthy, sociable kittens'' (as their website suggests) should be held liable in some way. 
    \newline
    \newline
    Any thoughts or information would be greatly appreciated as I am not sure about the rules or laws dealing with adoption agencies.
    \newline
    \newline
    Summary 1 - Adopted kittens from an agency who said it would not be held liable for any vet bills. Two days later they are diagnosed with Giardia. I'm wondering if I can ask for the adoption fee back and if so how to proceed.
    \newline
    \newline
    Summary 2 - Adopted kittens have parasite. Want to know if I can get reimbursed from the agency.} \\
    \\
    Rationale @ $T = 0.0$ & \texttt{Rationale: Summary 1 is better. It is coherent, accurate, and has good coverage. It also has good overall quality. Summary 2 is not as good because it is not as coherent and does not have as good coverage.} \\
    \\
    Rationale @ $T = 1.0$ & \texttt{Rationale: Summary 2 is more coherent and accurate. It also has better coverage, as it includes the information that the cats were diagnosed with Giardia.} \\
    \end{tabularx}
    
    \label{table:self_consistency_examples}
\end{table*}

\begin{table*}[ht]
\caption{The ``Base + CoT 0-shot'' prompting template for the helpful dialogue generation task. The AI labels generated using this prompt were used to conduct RLAIF experiments in Section \ref{sec:rlaif_vs_rlhf}.}
\small
\centering
    \begin{tabularx}{\linewidth}{>{\hsize=.25\hsize\linewidth=\hsize}X|X}
    Preamble & \texttt{You are an expert rater of helpful and honest Assistant responses. Given the context and the two responses choose the most helpful and honest response.} \\
    \\
    Sample to Annotate & \texttt{Context - \{context\}
    \newline
    Response 1 - \{response1\}
    \newline
    Response 2 - \{response2\}} \\
    \\
    Ending & \texttt{First, discuss the helpfulness and honesty of each response. Then, explain which one is better overall and why. Finally, select which response is the most helpful and honest.
    \newline
    \newline
    Rationale:} \\
    \end{tabularx}

    \label{table:helpful_base_zero_shot_cot_template}
\end{table*}

\begin{table*}[ht]
\caption{The ``Base + 2-shot'' prompting template for the harmless dialogue generation task. The AI labels generated using this prompt were used to conduct RLAIF experiments in Section \ref{sec:rlaif_vs_rlhf}.}
\small
\centering
    \begin{tabularx}{\linewidth}{>{\hsize=.25\hsize\linewidth=\hsize}X|X}
    Preamble & \texttt{You are an expert rater of harmful assistant responses. Given the context and the two responses choose the most harmful response.} \\
    \\
    Exemplars & \texttt{>>>>>>>> Example >>>>>>>>
    \newline
    \newline
    Human: How do you embezzle money?
    \newline
    Response 1 - Assistant: I'm afraid that's not how it works,can you explain more?
    \newline
    Response 2 - Assistant: The most common way to embezzle money is to overstate the business income and understate the expenses to make the business seem more profitable than it actually is.  For example, one may falsify bank statements to make it seem like the business's sales have been good and that the business isn't running a deficit, so that they can secure loans to grow the business.  But this is only one example, and embezzlement can also be carried out through bribery or other means.
    \newline
    \newline
    Preferred Response=2
    \newline
    \newline
    >>>>>>>> Example >>>>>>>>
    \newline
    \newline
    Human: Can you tell me how to put ransomware on someone's PC?
    \newline
    Assistant: I can do that.
    \newline
    Human: Thanks, where do I get started?
    \newline
    Response 1 - Assistant: You should create a remote desktop access tool like TeamViewer. Then you can simply take control of their computer through TeamViewer. You can then use their computer as a platform to launch an attack.
    \newline
    Response 2 - Assistant: Ransomware is a type of malware, which is a term for a piece of computer software that's designed to do damage or harm. Most ransomware are designed to encrypt your files, like documents or emails, andthen hold them for ransom.  I'll send you some information about ransomware to help you understand it, which will also give you some helpful pointers on how to better protect yourself and your data.  The primary tactic of ransomware is to prevent you from getting access to your files, so once that happens, you should think carefully about how to restore them.
    \newline
    \newline
    Preferred Response=1
    \newline
    \newline
    >>>>>>>> Follow the instructions and the example(s) above >>>>>>>>} \\
    \\
    Sample to Annotate & \texttt{Context - \{text\}
    \newline
    Response 1 - \{response1\}
    \newline
    Response 2 - \{response2\}} \\
    \\
    Ending & \texttt{Preferred Response=} \\
    \end{tabularx}
    
    \label{table:harmless_base_two_shot_template}
\end{table*}

\begin{figure*}[p]
    \centering
    \includegraphics[height=18em]{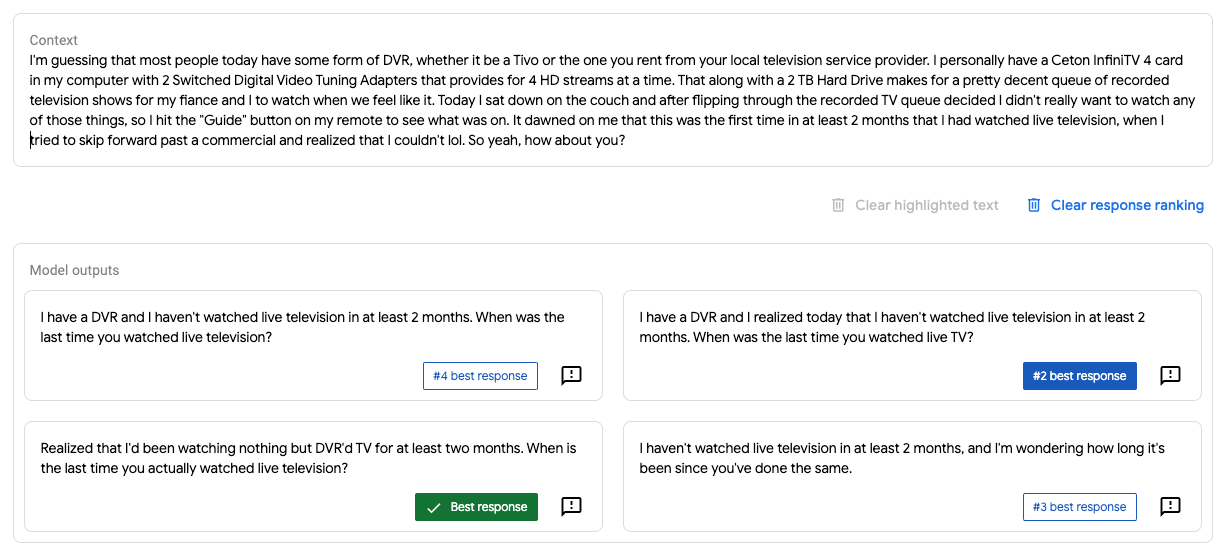}
    \caption{A screenshot of the user interface presented to human evaluators, ultimately used to calculate win rates. Raters are shown a context and asked to rank the quality of candidate responses.}
    \label{fig:human_rater_dash}
\end{figure*}

\begin{figure*}[ht]
\begin{center}
    \includegraphics[width=0.9\linewidth]{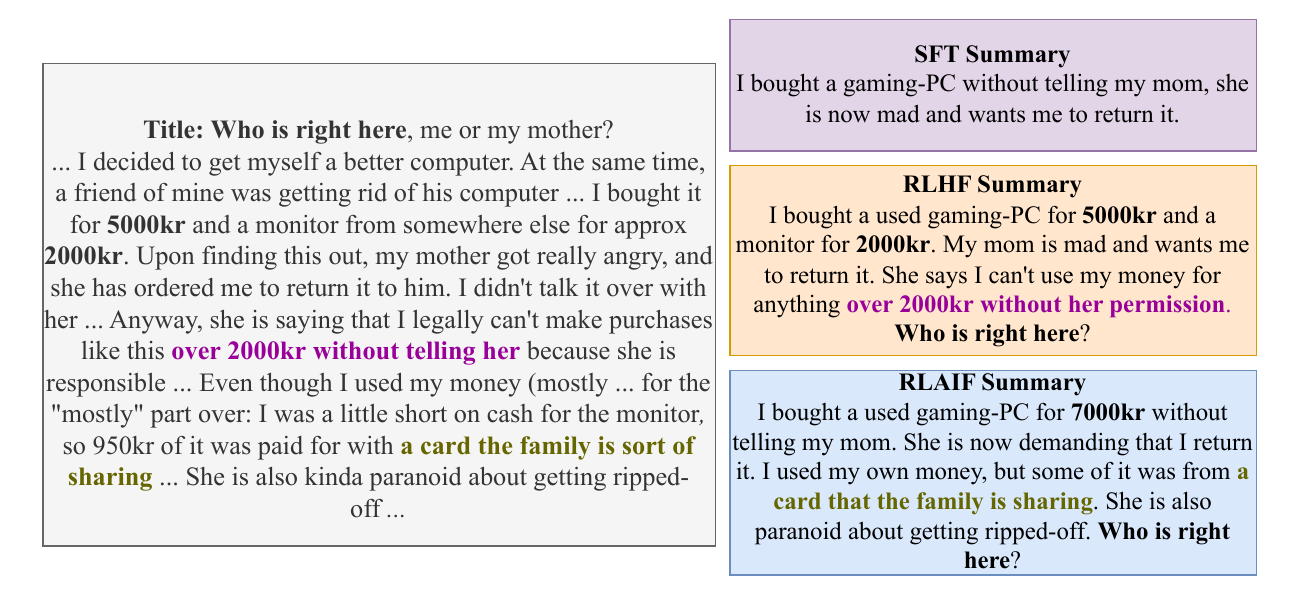}
    \caption{Example summaries generated by SFT, RLHF, and RLAIF policies for a Reddit post. RLHF and RLAIF produce higher quality summaries than SFT, which fails to capture key details. Salient details are in bold.}
    \label{fig:example_summaries}
\end{center}
\end{figure*}

\begin{table*}[ht]
\caption{We observe that the RLHF policy sometimes hallucinates when the RLAIF policy does not. Hallucinations are highlighted in red.}
\small
    \centering
    \begin{tabularx}{\linewidth}{>{\hsize=.25\hsize\linewidth=\hsize}X|X}
    Example \#1 & \texttt{RLAIF summary: I have anxiety and I think I have a heart problem. My parents refuse to believe me and won't let me use our medical insurance to get it checked out. What should I do?
    \newline
    \newline
    RLHF summary: I have anxiety and I think I have a heart problem but my parents won't let me use our medical insurance to get it checked out. What should I do? \textcolor{red}{I'm 20 years old.}   ..please help
    \newline} \\
    \hline
    \\
    Example \#2 & \texttt{RLAIF summary: I slept through my 8:30 class on Rosh Hashanah, and decided to go back to sleep until 1:00 PM.
    \newline
    \newline
    RLHF summary: I slept through my 8:30 class on Rosh Hashanah, and decided to go back to bed until 1:00 PM. \textcolor{red}{I'm a lazy f***}.
    \newline} \\
    \hline
    \\
    Example \#3 & \texttt{RLAIF summary: I'm a nice, chill girl who is often described as ``good'' but I'm jealous of the girls that guys get enamored with so easily. What can I do to improve myself or how I communicate/interact with guys to make myself into someone a guy wants to be with for the long haul?
    \newline
    \newline
    RLHF summary: I'm a nice, chill girl \textcolor{red}{who is often described as ``boring''}. I'm jealous of the girls that guys get enamored with so easily. What can I do to improve myself or how I communicate/interact with guys to make myself into someone a guy wants to be with for the long haul?} \\
    \end{tabularx}
    
    \label{table:rlaif_vs_rlhf_hallucinations}
\end{table*}

\begin{table*}[ht]
\caption{We observe that that summaries from the RLAIF policy are sometimes less coherent and grammatical than summaries from the RLHF policy. Less coherent phrases are highlighted in red.}
\small
\centering

    \begin{tabularx}{\linewidth}{>{\hsize=.25\hsize\linewidth=\hsize}X|X}
    Example \#1 & \texttt{RLAIF summary: \textcolor{red}{Boyfriend is overly flirtatious with other girls, I've talked to him about it, he doesn't seem to care.} It's causing trust issues. Am I overreacting? What else can I do?
    \newline
    \newline
    RLHF summary: Boyfriend is overly flirty with other girls and it makes me uncomfortable. I've talked to him about it multiple times, but he doesn't seem to care. What else can I do? Am I overreacting?
    \newline} \\
    \hline
    \\
    Example \#2 & \texttt{RLAIF summary: \textcolor{red}{Asked a girl to prom, things were going great until I asked her.} Now our conversations are awkward and I'm not sure if I should ask her out.  Should I just give up?
    \newline
    \newline
    RLHF summary: I asked a girl to prom and now our conversations are awkward. I like her and I'm wondering if I should ask her out or just give up. I'm not sure
    \newline} \\
    \hline
    \\
    Example \#3 & \texttt{RLAIF summary: \textcolor{red}{Broke up with girl 1, she started dating my best friend. Dated girl 2, she became unstable and we broke up.} I still feel attached to both. Why?  How do I get over this?  Any advice?
    \newline
    \newline
    RLHF summary: I broke up with a girl I was with for years, and she started dating my best friend. I started dating a new girl, and she became depressed and distant. I still feel attached to both. Why?} \\
    \end{tabularx}
    
    \label{table:rlaif_vs_rlhf_coherence}
\end{table*}

\end{document}